\documentclass[letterpaper]{comjnl}
\pdfoutput=1

\usepackage{booktabs} 
\usepackage[ruled]{algorithm2e}
\usepackage[nocompress]{cite}
\usepackage{multirow}
\usepackage{graphicx}
\usepackage{amsmath}
\usepackage{subfig}
\usepackage{amsthm}
\usepackage{url}
\usepackage{amsmath}

\begin{document}


\title[Work-Efficient Parallel Non-Maximum Suppression Kernels]{Work-Efficient Parallel Non-Maximum Suppression Kernels}

\author{David Oro}
\affiliation{Universitat Polit\`ecnica de Catalunya, C. Jordi Girona, 1-3, 08034, Barcelona, Spain}
\email{david.oro@upc.edu}

\author{Carles Fern\'andez}
\affiliation{Herta Security, C. Pau Claris, 165 4B, 08037, Barcelona, Spain}

\author{Xavier Martorell}
\affiliation{Universitat Polit\`ecnica de Catalunya, C. Jordi Girona, 1-3, 08034, Barcelona, Spain}

\author{Javier Hernando}
\affiliation{Universitat Polit\`ecnica de Catalunya, C. Jordi Girona, 1-3, 08034, Barcelona, Spain}

\shortauthors{D. Oro \emph{et al}}

\received{15 August 2019}
\revised{24 March 2020}
\accepted{10 July 2020}



\begin{abstract}
In the context of object detection, sliding-window classifiers and single-shot Convolutional Neural Network (CNN) meta-architectures typically yield multiple 
overlapping candidate windows with similar high scores around the true location of a particular object. Non-Maximum Suppression 
(NMS) is the process of selecting a single representative candidate within this cluster of detections, so as to obtain a unique 
detection per object appearing on a given picture. In this paper, we present a highly scalable NMS algorithm for embedded GPU 
architectures that is designed from scratch to handle workloads featuring thousands of simultaneous detections on a given picture. 
Our kernels are directly applicable to other sequential NMS algorithms such as FeatureNMS, Soft-NMS or AdaptiveNMS 
that share the inner workings of the classic greedy NMS method. The obtained performance results show that our parallel NMS algorithm 
is capable of clustering 1024 simultaneous detected objects per frame in roughly 1 ms on both Tegra X1 and Tegra X2 on-die GPUs, while 
taking 2 ms on Tegra K1. Furthermore, our proposed parallel greedy NMS algorithm yields a 14x-40x speed up when compared to 
state-of-the-art NMS methods that require learning a CNN from annotated data. 
\end{abstract}

\maketitle


\section{Introduction}

Recent advances in GPU computing performance have made the real-time execution of highly complex computer vision algorithms a reality. Applications 
including advanced driver-assistance systems (ADAS), autonomous driving, scene understanding, intelligent video analytics or face recognition, among others, 
usually leverage multithreaded data-parallel GPU architectures. In these environments, embedded computing is playing an increasingly important 
role due to the power consumption and thermal design point constraints imposed by small ubiquitous devices.

Typically, the most widely used object and event detection techniques for analyzing images and videos rely on the sliding window 
approach \cite{angelovareal,li2015convolutional} or single-shot CNN meta-architectures \cite{liu2016ssd}. Both types of classifiers yield multiple overlapping 
candidate windows with similar high scores around the true location of a particular object.

In this context, non-maximum suppression (NMS) is the process of selecting a single representative candidate within this cluster of detections, so as to obtain 
a unique detection per object appearing on a given picture. State-of-the-art CNN meta-architectures have also renewed the interest in applying fast NMS algorithms, 
as this process is mandatory after the forward pass of the network layers \cite{zhang2016joint}. As CNNs are inherently data parallel, they are usually 
built and fine-tuned using high-end discrete GPUs, which lately incorporate aggressive low-level hardware optimizations for speeding up both the training and 
inferencing processes \cite{volta18ipdps}. However, for the evaluation and deployment of such CNN models on real-world scenarios, embedded platforms featuring 
mobile programmable GPUs such as the NVIDIA Tegra \cite{nvidiaxaviersite} are quickly gaining traction.

Modern system-on-chip heterogeneous platforms feature low-power multicore out-of-order Armv8 CPU cores combined with general-purpose GPUs, which consume a large 
part of the die area. These embedded GPUs are quickly closing the performance gap with high-end discrete GPUs, and they are now powerful enough for handling 
massively parallel CUDA and OpenCL kernels. Starting from the Tegra K1, all successive NVIDIA SoCs (e.g. Tegra X1, X2,  \emph{Xavier}, \emph{Orin}) implement scaled-down 
versions of the GPU architectures included in the discrete GPU variants (i.e. \emph{Kepler}, \emph{Maxwell}, \emph{Pascal}, \emph{Volta}, \emph{Ampere} \cite{nvidiaarchs}), 
thus opening the door for scheduling and executing exactly the same CUDA kernels used in supercomputers and data centers, but at a substantially lower power budget. Unfortunately, 
the NMS process is still sequentially executed on CPUs and thus cannot exploit the vast amount of computing resources available on general-purpose embedded GPUs.

For real-time computer vision applications analyzing large amounts of simultaneous objects, such as facial recognition in large crowds or autonomous vehicles featuring 
L5 capabilities \cite{litman2017autonomous}, a data-parallel GPU kernel that overcomes the latency constraints imposed by the data dependencies of serial 
NMS implementations is becoming increasingly necessary. More particularly, as the input resolution and the amount of camera video streams to be analyzed by a low-power 
SoC keeps growing, it will be unfeasible to compute the NMS process on a given CPU core sequentially for two main reasons: (i)  the overwhelming number of objects detected, 
and (ii) the waste in DMA, I/O, power and memory bandwidth resources as a result of unwanted GPU-to-CPU and CPU-to-GPU transfers. These memory transfers are a must, should 
all remaining stages of the smart video processing pipeline be fully offloaded to the GPU. The problem is further compounded when it is required to track and 
recognize objects on a frame per frame basis for safety or security reasons.

As we showed in a previous study \cite{oro2016work}, it is possible to exploit the usage of a boolean adjacency matrix to avoid data dependencies when performing 
clustering and thus solve the problem of NMS in parallel. Our idea even inspired the design of customized circuits using transistors \cite{shi2019fast} to rapidly 
solve NMS computations in a power efficient manner.

In this paper, we conduct an in-depth study of a pair of CUDA \emph{map/reduce} kernels to solve the problem of NMS in a work-efficient manner by relying on such boolean 
adjacency matrix. More particularly, the main contributions of this paper are the following. We demonstrate that the theoretical asymptotic time complexity of our parallel NMS 
algorithm is linear. We analyze the scalability of our proposed parallel NMS algorithm by providing an exhaustive experimentation on several platforms (i.e. multiple NVIDIA 
Tegra SoCs, discrete NVIDIA GeForce GTX 1060 and Tesla T4) with increased core counts, and improved memory bandwidth over both a picture and a challenging video dataset. The 
selected workloads feature large amounts of simultaneous objects to maximize parallelism and saturate the underlying hardware resources. Finally, we include the proof of 
correctness of the kernels constituting our parallel NMS algorithm.

The remaining part of the paper is structured as follows. Section \ref{sec:rwork} includes related work describing other NMS approaches found in the literature. Section 
\ref{sec:impl} presents the inner workings of our proposed kernels for solving in parallel the NMS problem. Section \ref{sec:complexity} studies the time complexity of our 
parallel NMS algorithm. Section \ref{sec:methodology} introduces the evaluation methodology used for experimentation and to study the scalability of the parallel 
NMS algorithm. Section 6 describes the obtained performance results. Section 7 shows the proof of correctness. Finally, in Section 8 conclusions are drawn.

\section{Related Work}
\label{sec:rwork}

Traditional frameworks that compute hand-crafted features by evaluating a classifier derived from boosting machine learning techniques \cite{schapire2003boosting} or 
SVMs \cite{cortes1995support} generate multiple detections surrounding the ground-truth location of a pre-trained object class. State-of-the-art CNN architectures 
also require fusing multiple overlapping detections.

\begin{figure}[t]
  \centering
  \includegraphics[scale=1.2]{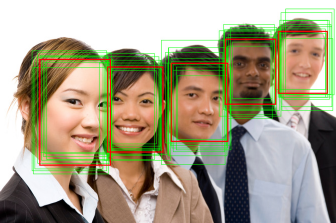}
  \caption{\label{fig:nmsab} Visualization of the NMS process for a CNN-based face classifier. Pre-NMS (green boxes). Post-NMS (red box).}
\end{figure}

As Figure \ref{fig:nmsab} depicts, this fusion is an important step in the context of face detection. Usually, the output of each detected window 
is a score derived from the last layer of the CNN. More particularly, this score represents a measure of the likelihood that the region enclosed by the window contains 
the object for which the CNN classifier has been trained. The score is thus degraded as the location and scale of the sliding window containing the object varies. 
As a result, the maximum score is obtained at the precise location and window dimensions, corresponding to the local maximum of the response function used by the CNN.

The goal of NMS is to extract a good, single representative from each set of clustered candidate object detections. Therefore, NMS resembles a classic clustering 
problem, and typically relies on two basic operations: (i) identifying the cluster to which each detection belongs, and (ii) finding a representative for each cluster.

Assuming rectangular bounding boxes, the positive output of a given sliding-window classifier yields a tuple $\{x, y, w, h, s\}$, namely 2D coordinates $(x,y)$, window 
width and height $w \times h$, and a score $s$ for a detection $d \in D$, in which $D$ is the set containing all detected objects. This NMS approach is usually implemented 
as a greedy iterative process, and involves defining a measure of similarity between windows while setting a threshold $\theta$ for window suppression.

Recent works \cite{bodla2017soft,liu2019adaptive,salscheider2020featurenms} commonly rely on the abovementioned greedy NMS technique, as it still obtains 
the best accuracy when average precision (AP) is used as an evaluation metric, and does not require dedicated training. These post-processing methods essentially 
find the window with the maximum score, and then reject the remaining candidate windows if they have an intersection over union (IoU) larger than a learned threshold. 
However, in such works NMS parallelization remains unaddressed. 

Another common NMS approach is to employ optimized versions of clustering algorithms, particularly \emph{k-means} \cite{shalom2008efficient} or 
\emph{mean shift} \cite{li2009mean}. Unfortunately, \emph{k-means} requires a predetermined number of clusters, which is unknown and difficult to estimate beforehand; 
and additionally only identifies convex clusters, so it cannot handle very non-linear data. On the other hand, \emph{mean shift} is computationally intensive and often 
struggles with data outliers. Combining both methods may solve many of these problems in practice, but their iterative nature makes them difficult to parallelize, and 
highly uncompetitive from a latency perspective.

The \emph{affinity propagation} clustering algorithm \cite{rothe2014nms} overcomes the shortcomings derived from hard-coded thresholds of greedy NMS methods. Nevertheless, 
this proposal is unworkable for real-time applications, as the authors report a latency of 1000 ms to cluster 250 candidate windows on an unreported computing platform.

Although it is possible to train a neural network for solving the NMS problem, recent works prove that the accuracy improvements are minimal when compared 
to traditional greedy NMS. Hosang et al. \cite{hosang2017learning} designed a CNN architecture (\emph{GNet}) to directly learn and solve the problem of 
NMS without relying on an isolated  human-designed algorithm for performing the clustering of detections. The authors report that \emph{GNet} outperforms by 1.6\% the 
traditional greedy NMS algorithm in terms of AP on the PETS \cite{ellis2010pets2010} and MS COCO \cite{lin2014microsoft} datasets. Unfortunately, it requires 
huge amounts of training data, and most importantly, the authors report 14 ms of latency just for the NMS network. The evaluation was performed 
on a power-hungry NVIDIA Tesla K40M when clustering pictures from the prior datasets, which contain only 67.3 detections per image on average.

Qiu et al.~\cite{qiu2019graph} proposed NMSNet, a network that relies on graph convolutions and self attention. This proposal improves the 
accuracy of greedy NMS only between 1\% and 2\% in terms of mAP for some object classes within the PASCAL VOC 2007~\cite{everingham2010pascal} dataset. However, the authors 
report worse mAP than the classic greedy NMS in classes such as chairs, bottles, TVs or birds. Also, the reported latency of NMSNet was 40 ms on an unspecified platform. 
Therefore, this network could hardly be used by real-time applications immediately after the output of an object detector, as by itself consumes the 40 ms required for 
real-time processing.

Another recent proposal introduced by Song et al.~\cite{song2019improved} relies on a \emph{harmony search} (HS) algorithm for solving the problem of NMS. Again, the reported 
mAP of this algorithm only improves the accuracy at most 1\% for the PASCAL VOC 2007 and MS COCO datasets when compared to the traditional NMS. For 9 of the 20 object classes of 
the PASCAL VOC 2007 dataset, the obtained accuracy was lower than the classic NMS method. The authors did not report the execution time of HS-NMS, but as it internally relies 
on the classic NMS plus sorting and harmony search heuristics, its latency should be definitely higher than the greedy NMS.

Other proposals, such as FeatureNMS~\cite{salscheider2020featurenms}, extend the classic NMS by computing the $L^{2}$ distance of feature embeddings between bounding boxes 
when the IoU is in a range that makes difficult to make a decision. This approach improves roughly 2\% the AP in the CrowdHuman~\cite{shao2018crowdhuman} dataset when compared 
to the classic greedy NMS. Other minimal modifications in the classic greedy NMS are considered in Soft-NMS~\cite{bodla2017soft}, which updates the detection scores by rescaling 
them using a linear or Gaussian function, and keeping the rest of the algorithm equal. This minor change improves the accuracy in terms of mAP 1.7\% in the PASCAL VOC 2007, and 
roughly 1\% in the MS COCO dataset. The computational complexity of the algorithm remains $\mathcal{O}(n^{2})$, as in the generic greedy NMS. Similarly, AdaptiveNMS~\cite{liu2019adaptive} 
also extends the classic greedy NMS method. Internally, this proposal adaptively suppresses detections, and later scales their NMS threshold according to their detection densities, 
which are learnt using a sub-network that must be embedded in the preceding CNN-based detector. The authors report at most a 1.6\% improvement in terms of AP for the CrowdHuman 
dataset, but they compare their approach against the classic NMS using different working points that cannot be compared. For this reason, they rely instead on the miss rate on 
false positives per image (denoted as MR$^{-2}$). However, AdaptiveNMS only reduces a 2.62\% the MR$^{-2}$ when compared to greedy NMS, albeit at the cost of modifying the CNN 
used for performing object localization.

\begin{figure}[t]
\centering
\includegraphics[scale=0.21]{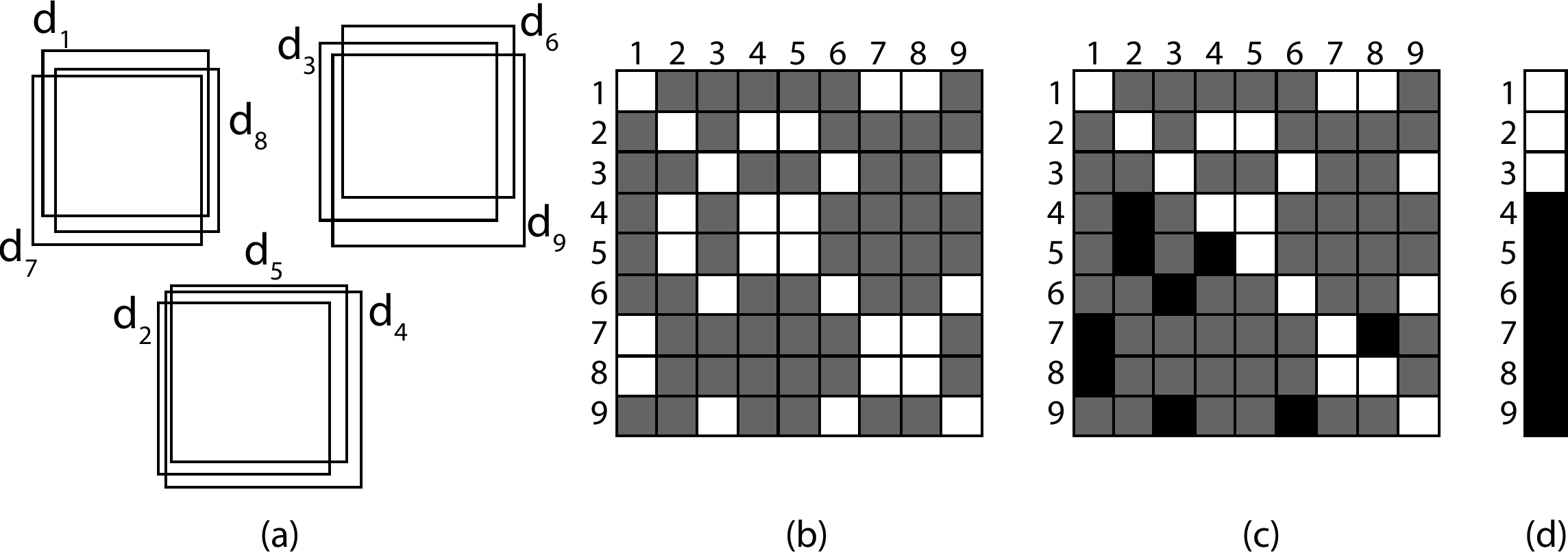}
\caption{\label{fig:mapkernel} Visualization of our GPU-NMS proposal: 
(a) example candidates generated by a detector (3 objects, 9 detections); 
boolean matrix after (b) clustering and (c) cancellation of 
non-representatives; and (d) result after AND reduction.}
\end{figure}

In view of the recent works, the classic hand-crafted greedy NMS algorithm still remains competitive and effective in terms of mAP, as it outperforms state-of-the-art neural network-based 
approaches in several object classes. Therefore, it is unclear how NMS methods based on neural networks are going to evolve to support image workloads featuring thousands of 
simultaneous detections, while clustering them in a few milliseconds to efficiently target real-time applications. Moreover, these NMS networks should also be rearchitected for embedded 
GPU platforms to enable the use cases discussed earlier, as they were evaluated on high-end discrete GPUs. On the other hand, minor variations of the classic NMS such 
as FeatureNMS, Soft-NMS or AdaptiveNMS share the same core and inner workings of the greedy algorithm. As a result, the parallelization pattern studied in this paper is also 
applicable to these recent serial-based NMS proposals, as it only involves minor modifications during the merging process of detections.

Finally, Shi et al.\cite{shi2019fast} implemented a power-efficient hardware microarchitecture for NMS directly in silicon (28 nanometer). This low-level chip design is inspired by 
our previous work \cite{oro2016work}, and as such is cited by the authors. By following this strategy, the designed chip is capable of merging 1000 candidate boxes in 12.79 $\mu s$ by consuming only 6.142 mW. These results further reinforce the validity of our algorithm. 

Besides of customized CNN architectures that learn the NMS process with annotated data, to the best of our knowledge, there is no other way to tackle the NMS problem by means of a 
lightweight greedy data-parallel algorithm specifically tailored to GPUs.

\section{Proposed Parallel Algorithm}
\label{sec:impl}

In order to exploit the underlying architecture of general-purpose embedded GPUs, an NMS kernel must expose a parallelization pattern in which each computing thread 
independently evaluates the overlapping between two given bounding boxes. The idea is to avoid, to the maximum extent possible, data dependencies that serialize computations, and 
thus overcome the limitations in scalability derived from the traditionally iterative clustering process. Our proposal addresses this issue by adopting a \emph{map/reduce} 
parallelization pattern which uses a boolean matrix both to encode unsorted candidate object detections and to compute their cluster representatives.

\begin{algorithm}[t]
\DontPrintSemicolon
\SetAlgoNoLine
\KwData{Matrix $B$ and vector $D$}
\caption{\label{alg:map}\textsc{MapKernel}\label{IR}}
\Begin{
$i \leftarrow$ \footnotesize\texttt{blockIdx.x * blockDim.x + threadIdx.x}\normalsize\;
$j \leftarrow$ \footnotesize\texttt{blockIdx.y * blockDim.y + threadIdx.y}\normalsize\;
\vspace{0.04in}
\If{\footnotesize$D[i].s < D[j].s$}{
\vspace{0.025in}
\normalsize$a \leftarrow$ \footnotesize $(D[j].z + 1) * (D[j].z + 1)$\;
\vspace{0.025in}
\normalsize$w \leftarrow$ \footnotesize $max(0, min(D[i].x + D[i].z, D[j].x + D[j].z)$\;
\vspace{0.025in} 
\hspace{0.3in} $ - max(D[i].x, D[j].x) + 1)$\;
\vspace{0.025in}
\normalsize$h \leftarrow$ \footnotesize $max(0, min(D[i].y + D[i].z, D[j].y + D[j].z)$\; 
\vspace{0.025in}
\hspace{0.3in} $ - max(D[i].y, D[j].y) + 1)$\;
\vspace{0.025in}
$B[i * D_{max} + j] \leftarrow (\frac{w*h}{a} < \theta) \wedge D[j].z \neq 0$\;
}
}
\end{algorithm}

\begin{algorithm}[t]
\DontPrintSemicolon
\SetAlgoNoLine
\KwData{Matrix $B$, value $k$, and vector $V$ of size $D_{max}$}
\caption{\label{alg:reduce}\textsc{ReduceKernel}\label{IR}}
\Begin{
$i \leftarrow $ \footnotesize\texttt{blockIdx.x}\normalsize\;
$j \leftarrow i * D_{max} +$ \footnotesize\texttt{threadIdx.x}\normalsize\;
$n \leftarrow D_{max} / k$\;
\vspace{0.025in}
$V[i] \gets $ \footnotesize\texttt{\_\_syncthreads\_and}\normalsize$(B[j])$\;
\vspace{0.025in}
\For{$1$ \KwTo $k - 1$}{
$j \gets j + n$\;
$V[i] \gets $ \footnotesize\texttt{\_\_syncthreads\_and}\normalsize$(V[i]$ \&\& $B[j])$\;
}
}
\end{algorithm}

Figure \ref{fig:mapkernel} depicts a toy example of the proposed algorithm, in which an image frame contains
three objects, three window clusters and nine detections. Our matrix encodes the relationship 
among all detections, initially assuming that all are possible cluster representatives (all matrix values are set to one after 
memory allocation). Therefore, a white color matrix element in Figure \ref{fig:mapkernel} represents updated elements that were 
reset to boolean \emph{true} values (also, implemented using ones) while an element depicted in gray color means that the element 
has not been updated (original values that were set to one during initialization). Finally, an element depicted in black color 
means that it has been reconsidered as non-representative and replaced by a zero (boolean \emph{false} value).

\begin{figure*}[!t]
\centering
\includegraphics[scale=0.8]{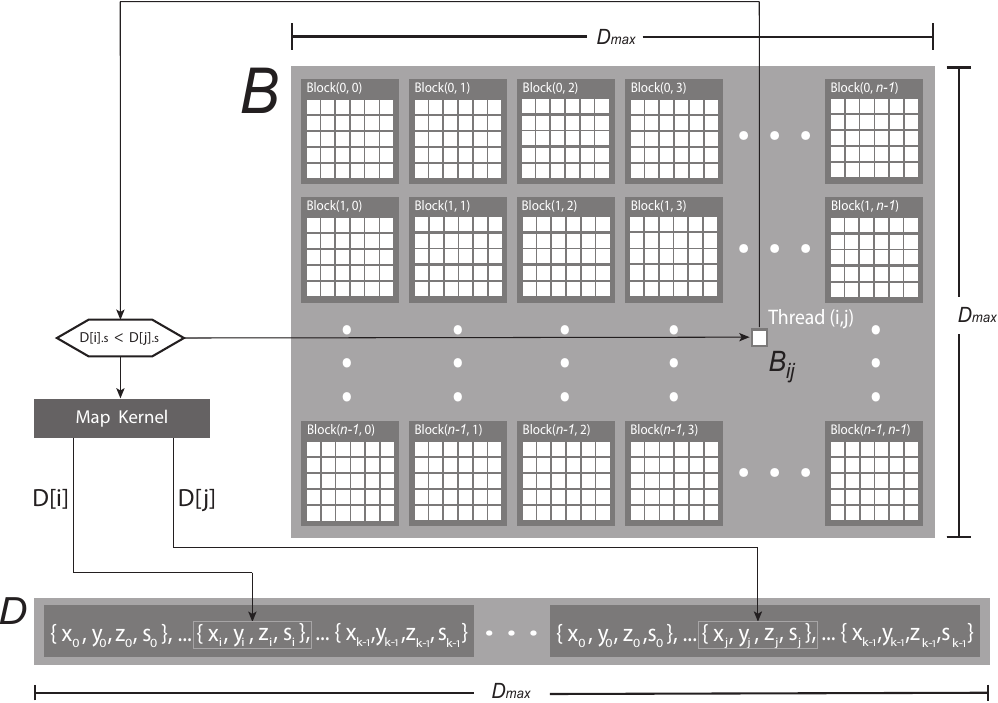}
\caption{\label{fig:mapkernelcuda} Representation of the NMS \emph{map} kernel using the CUDA programming model notation.}
\end{figure*}

Firstly, we decide that two windows $d_i$ and $d_j$ belong to the same cluster if their 
areas are overlapped beyond a given threshold; otherwise, a zero will be placed in the matrix coordinates 
$(d_i,d_j)$ and $(d_j,d_i)$. Secondly, we evaluate the non-zero values of each row, and again place zeroes
if the row-indexed detection ($d_i$) is strictly smaller than the column-indexed one ($d_j$), thus discarding
$d_i$ as the cluster representative (blacked out in Figure \ref{fig:mapkernel}). Finally, a horizontal AND reduction 
will preserve a single representative per cluster, thus completing the NMS.

Formalizing this process, let $D$ be the set of detection windows and $C$ the set of clusters for a 
given frame, with $C\subseteq D$. We build a boolean matrix $B$ of size $D_{max}\times D_{max}$, being $D_{max}$
an upper limit of the number of windows possibly generated by the detector at any frame. Let $A(\cdot)$ be 
an operator that returns the area of a window. Given an overlapping threshold $\theta\in\left[0,1\right]$, 
two candidate windows $d_i$ and $d_j$ are assigned to the same cluster if $A(d_i\cap d_j)/A(d_i)\geq \theta$.
Candidates within a cluster are discarded as representatives if $A(d_i)<A(d_j)$. The clipping process is 
performed in parallel independently for each detection $d\in D$ by a given computing thread. Since there 
are no data dependencies among detections, this mapping strategy scales properly as the amount of GPU cores 
is increased.

The only required parameters for this algorithm are the overlapping threshold $\theta$ and the maximum number 
of possible candidates that can be generated by the detector $D_{max}$. Although this last constraint may 
initially seem to be an important drawback, in general conservative values for $D_{max}$ turn to be very 
relaxed constraints. As an example, it is common that face detection models have a minimum face resolution 
of $24\times 24$ pixels. In that case, the worst case scenario for a HD frame of $1920\times 1080$ pixels 
would be a tiling of $80\times 45$ faces of that size, yielding a matrix $B$ of $3600^2$ elements.

Internally, the \emph{map} kernel (see the CUDA pseudocode in Algorithm \ref{alg:map}) must first compute the area $a$ 
to effectively perform the overlapping test for each pair of detections $d_i, d_j \in D$. With the aim of 
preserving simplicity, we assume equal width and height for the bounding boxes. Therefore, each detection is 
redefined as a $\{x, y, z, s\}$ tuple, in which $z=w=h$.

Moreover, each $d_i$ detection included in the $D$ set can now be formulated as a vector of tuples stored in any arbitrary 
order (i.e. unsorted), as it is described in the following Equation \ref{eq:dvector}:

\begin{equation}
\label{eq:dvector}
D = \langle \{ x_0, y_0, z_0\, s_0 \}, ... \{ x_{n-1}, y_{n-1}, z_{n-1}, s_{n-1} \} \rangle
\end{equation}

\bigbreak

The process of constructing the boolean $B$ matrix is implemented by relying on a classic 2D parallelization 
pattern (see Figure \ref{fig:mapkernelcuda}) in which each thread within a CUDA block updates a matrix element. Initially, all values 
of the $B$ matrix are set to 1 by calling the \small\texttt{cudaMemset()} \normalsize function before executing the \emph{map} kernel to 
ensure the correctness of the algorithm. At this point, considering that the $D$ vector constitutes a read-only input, it is possible to 
simultaneously access its values from multiple threads in the \emph{map} kernel without relying on synchronization primitives for implementing 
the pre-clustering of detections according to their scores. Therefore, in Algorithm \ref{alg:map}, the \textbf{if} statement 
checks $D[i].s < D[j].s$ to ensure that a given block thread only updates a boolean element $B_{ij} \in B$ corresponding to a 
detection $d_i \in D$ when its score is higher than the one of $d_j \in D$.

An additional second conditional check is encoded as a boolean expression in the kernel $(\frac{w*h}{a} < \theta) \wedge D[j].z \neq 0$ for 
re-tagging the $B_{ij}$ element either as cluster candidate or non-candidate depending on the area overlap between $d_i$ and $d_j$. 
The overlap is computed by means of $max$ and $min$ functions using as an input the $x$ and $y$ coordinates and the height and width of each 
$(d_i,d_j)$ pair, whose dimensions depend only on its $z$ component ($z \times z$). It also discards empty values of the $D$ vector, as its 
size is allocated to an upper limit $D_{max}$ that may not exactly match the total number of detections. This is ensured by the $D[j].z \neq 0$ 
condition, which requires, as a prerequisite, all elements of the $D$ vector initialized to 0 with \small\texttt{cudaMemset()} \normalsize before 
storing on it the input detections to be merged by the NMS algorithm. Finally, the $B_{ij}$ element is updated only if the overlapping 
ratio $\frac{w*h}{a}$ exceeds the hand-crafted $\theta$ threshold considered in the NMS process, as it happens with the conventional serial greedy NMS.

Once the boolean matrix $B$ has been computed, it is required to call a \emph{reduce} kernel (see Algorithm \ref{alg:reduce}) for 
selecting the optimal candidate from each row as it is depicted in Figure \ref{fig:mapkernel}. This task is performed using 
AND operations in parallel for each row of $B$ and can be implemented in a CUDA kernel by means of \small\texttt{\_\_syncthreads\_and(cond)}. \normalsize This 
directive returns 1 only if the \texttt{cond} predicate evaluates to true for all threads of the 
CUDA block, and is directly translated to the hardware-accelerated \small\texttt{BAR.RED.AND} \normalsize 
assembly instruction. Therefore, it is possible to split the AND reductions of $B$ by creating 
$D_{max} / k$ partitions per row, and then assigning each partition to a given thread block. As a proof of concept, Figure \ref{fig:reducekernelcuda} 
shows how the \emph{reduce} kernel creates $n$ CUDA blocks per partition (highlighted using dashed rectangles), thus precisely matching the 
number of rows of boolean matrix $B$.

Under this parallelization pattern, threads are synchronized, and simultaneously reduce the boolean values 
stored in the CUDA block of each row partition by relying on the abovementioned directive \small\texttt{\_\_syncthreads\_and(V[i])}\normalsize. 
Partial row reductions of $B$ are later stored in $V[i]$, a boolean vector $V$ of size $D_{max}$. Since we are dealing with a square matrix, it is 
required to call $k$ times the reduction directive within the kernel assuming a CUDA block of size $D_{max} / k$ and a grid size 
equal to the size of the $D$ input set detections.

Consequently, after the first reduction ends, a \textbf{for} loop wraps up the remaining $k-1$ reductions by 
calling \small\texttt{\_\_syncthreads\_and() }\normalsize using as an input the partial AND reduction values 
stored in $V[i]$. An additional AND operation with the current thread block completes the operation (coded as $B[j]$ \&\& $V[i]$), 
effectively computing the correct aggregated reduction. On the other hand, parameter $k$ must be experimentally determined so that 
the GPU achieves the highest occupancy and yields the minimum latency.

When the execution of the \emph{reduce} kernel finishes, the boolean values stored in the $V$ vector encode enough information to conclude  
the NMS process. Referring again to the $D$ vector of detections described in Equation \ref{eq:dvector}, the bitmask information 
contained in $V$ is used for indexing which detections are survivors, and which ones must be discarded. This final post-processing step 
is formalized as follows in Equation \ref{eq:dvectornms}:

\begin{equation}
\label{eq:dvectornms} 
D_{\tiny\textsf{NMS}\normalsize} = \langle \bigcup_{i=1}^{D_{max}} \left(d_i \wedge v_i \right) \rangle \enspace \textrm{where} \enspace 
d_i \in D \enspace \textrm{,} \enspace v_i \in V
\end{equation}

After having bitmasked all $D$ vector elements with the boolean values stored in $V$, the $D_{\tiny\textsf{NMS}\normalsize}$ vector 
elements correspond to the merged set of detections (i.e. best representative of each cluster of detections). At this point, the 
execution of the parallel NMS algorithm is fully completed.

\section{Algorithm Complexity}
\label{sec:complexity}

One of the goals of this study is to determine the time complexity of the kernels implementing the parallel NMS algorithm. In view 
of our kernels are meant to be executed on NVIDIA GPUs, which are mainly multithreaded data-parallel and throughput-oriented processors, 
their asymptotic time complexity must be determined using an idealized model of a parallel machine. In order to do so, we selected the 
parallel random-access machine (PRAM) 
model \cite{karp1988survey} \cite{fortune1978parallelism}, which is mainly an idealized shared address space 
parallel computer equipped with $p$ processors lacking resource contention mechanisms. The generic PRAM model assumes 
that a given $p$ processor has random access in unit time to any address of the external memory.

\begin{figure*}[!t]
\centering
\includegraphics[scale=1.2]{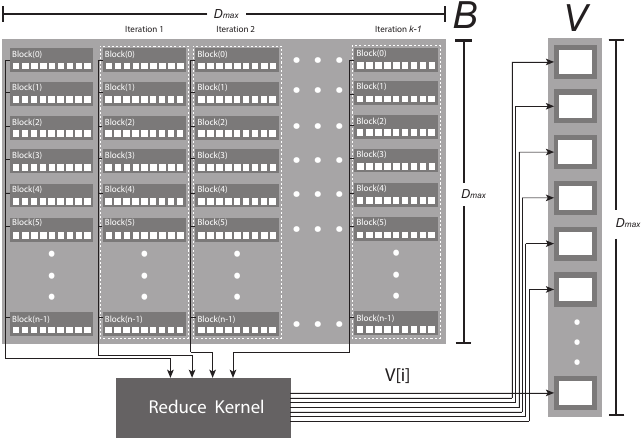}
\caption{\label{fig:reducekernelcuda} Illustration of the NMS \emph{reduce} kernel showing partitions and CUDA blocks.}
\end{figure*}

 \begin{table*}
  \renewcommand{\arraystretch}{1.3}
  \caption{Selected embedded NVIDIA Tegra platforms.}
  \centering
  \footnotesize
  \begin{tabular}{| c | c | c | c | c | c |}
  \hline
  \textbf{Tegra SoC} & \textbf{CPU Cores} & \textbf{CUDA Cores} & \textbf{GPU Architecture} & \textbf{GPU Clock Frequency} \\
  \hline
  T124 & (4+1) Arm Cortex A15 & 192 & \emph{Kepler} (GK20A) [\texttt{sm\_32}] & 72 MHz - 852 MHz \\
  T210 & (4) Arm Cortex A57 & 256 & \emph{Maxwell} (GM20B) [\texttt{sm\_53}] & 76.8 MHz - 998.4 MHz \\
  T186 & (2) Denver2 + (4) Arm Cortex A57 & 256 & \emph{Pascal} (GP10B) [\texttt{sm\_62}] & 114.75 MHz - 1.3 GHz \\
  \hline
  \end{tabular}
  \normalsize
  \label{tab:jetsonplatforms}
 \end{table*}

However, in order to realistically match a modern GPU, we selected a PRAM with concurrent read and concurrent 
write (CRCW) capabilities to the shared memory address space, as these particular read/write conflicts are usually managed 
by the programmer using the synchronization and atomic primitives offered by the CUDA programming model. It should be 
noted that the shared memory address space referred in the PRAM model corresponds to the \emph{global memory} GPU address 
space (external DRAM), and not to the on-die \emph{shared memory} included in the simultaneous multiprocessors (SMs) used 
in the standard CUDA-enabled GPU architecture.

By following these assumptions, starting from an input $D$ vector of detections of size $n = D_{max}$, Algorithm \ref{alg:map} 
builds a matrix of size $D_{max} \times D_{max}$. Therefore, Algorithm \ref{alg:map} must populate $n^2$ elements of 
matrix $B$. As a result of this, the space complexity of our method is $\Theta(n^2)$. Considering that the PRAM model allows 
concurrent read accesses at zero cost to vector $D$, all operations executed and 
enclosed by the statement \textbf{if} $D[i].s < D[j].s$ are computed in $\Theta(1)$ time on each processor $p$. Similarly, 
all store memory operations writing $B$ elements are also computed concurrently, even in the case when $i = j$, in which 
$B$ values are overwritten albeit guaranteeing the clustering correctness. Consequently, the asymptotic time complexity of 
Algorithm \ref{alg:map} can be estimated as follows:

\begin{equation}
\label{eq:tmap}
T_{\tiny\textsf{MAP}\normalsize} = \Theta\left(\frac{n^{2}}{p}\right) \quad \textrm{where} \quad p = O(n)
\end{equation}

\bigbreak

The $T_{\tiny\textsf{MAP}\normalsize}$ asymptotic cost shown in Equation \ref{eq:tmap} treats the $p$ amount of 
processors as another variable in our analysis, in which $p$ is expressed as a cost function of input $n$. This fact means 
that when Algorithm \ref{alg:map} is executed on a PRAM machine that has exactly enough $p$ processors to exploit the 
maximum concurrency with $n$ parallel operations, it behaves as a linear algorithm (i.e. $T_{\tiny\textsf{MAP}\normalsize} = \Theta\left(n\right)$).

On the other hand, assuming near-zero cost in thread synchronization primitives, the reduction kernel listed in Algorithm 
\ref{alg:reduce} performs $\Theta(k)$ operations (i.e. internal \textbf{for} loop) on each processor $p$. In this case, $k$ 
corresponds to the column partitions in matrix $B$ shown in Figure \ref{fig:reducekernelcuda}, whereas $n$ denotes 
the size of vector $V$, which precisely matches the row size of matrix $B$.

\begin{equation}
\label{eq:treduce}
T_{\tiny\textsf{REDUCE}\normalsize} = \Theta\left(\frac{nk}{p}\right) \quad \textrm{where} \quad p = O(n)
\end{equation}

\bigbreak

The asymptotic time complexity of the reduction kernel is summarized in Equation \ref{eq:treduce}, and considering again 
a PRAM machine with $p$ processors capable of computing $n$ concurrent operations, $T_{\tiny\textsf{REDUCE}\normalsize}$ 
could be simply approximated as $\Theta\left(k\right)$.

Therefore, when adding the asymptotic time complexities of both \emph{map/reduce} kernels (Equation \ref{eq:tnms}), 
we can conclude that given enough processors, the proposed parallel NMS method approximately behaves as a linear algorithm:

\small
\begin{equation}
\label{eq:tnms}
T_{\tiny\textsf{NMS}\normalsize} = T_{\tiny\textsf{MAP}\normalsize} + T_{\tiny\textsf{REDUCE}\normalsize} = \Theta\left(\frac{n^2 + nk}{p}\right) \simeq \Theta\left(n + k\right)  
\end{equation}
\normalsize

Although this time complexity analysis may be considered somewhat simplistic, as it does not take into account the capabilities of 
SM warp schedulers to issue and execute several warps concurrently nor considers the intrinsic low-level delays of the on-die GPU 
interconnect, it roughly highlights the computational burden of the two CUDA kernels. As a result, further experimentation and profiling 
on real platforms is required to determine the optimal $k$ partition parameter, and also to study the parallel NMS scalability 
as both the CUDA core count and the amount of objects to cluster is increased.

\section{Evaluation Methodology}
\label{sec:methodology}

Several evaluations were conducted to precisely determine the latency of our proposed parallel NMS method. Since the main 
use case of such algorithm is to enable the clustering of a high number of simultaneous detections per frame on low-power 
embedded devices, the platforms selected for running the \emph{map/reduce} kernels were the NVIDIA family of Tegra SoCs, 
which enable the scheduling, execution, profiling and debuggability of CUDA kernels with minimal efforts.

Tables \ref{tab:jetsonplatforms} and \ref{tab:jetsondistros} summarize the three boards used for running the 
experiments, as well as the version of the flashed NVIDIA JetPack image, which includes the required \emph{Linux for Tegra} (L4T) 
OS distribution \cite{nvidial4t}. The source code of our kernels was compiled with NVIDIA's \texttt{nvcc} compiler using the \texttt{-O3} flag 
on each Tegra platform. This architectural diversity helped us on quantifying the scalability of the parallel algorithms 
implemented in our NMS kernels. More particularly, we were interested in benchmarking the NMS kernels across several generations of 
the Tegra SoCs featuring on-die GPUs with higher CUDA core counts, improved microarchitectures, memory bandwidth, bus width, 
and clock frequencies. 

In order to do so, the clock rate of the GPU was manually set before each kernel execution by means of the Linux kernel 
pseudo filesystem device interface (\texttt{sysfs}). The path for setting the GPU frequency differed depending on the 
Tegra SoC model included in the embedded board (see Table \ref{tab:tegrafs}). This low-level fine tuning opened us the door 
for comparing the throughput-oriented GPU architecture with the latency-oriented CPU cores when running both the conventional 
serial CPU-based NMS and the GPU-based parallel NMS at multiple clock frequencies.

The dynamic voltage and frequency scaling (DVFS) mechanisms of the CPUs were also disabled on purpose, and their corresponding clock 
frequencies manually set up for conducting a fair embedded CPU versus GPU comparison. Unfortunately, only the clustered multicore 
CPU design of the Tegra K1 SoC (T124) offered us the possibility of having access to an architecture packing high-performance  
out-of-order CPU cores with a low-performing and low-power CPU core (the so-called fifth \emph{companion core}). The Tegra T210 SoC 
included in the Jetson TX1 board features only four high-performance Arm Cortex A57 CPU cores. Even though that the SoC die includes 
four additional in-order Arm Cortex A53 CPU cores, they were disabled on purpose during the manufacturing process. Therefore, this SoC does not provide 
OS-level access to a low-power cluster set of cores for reducing the power consumption when running background tasks, as opposed to other chips 
implementing Arm's big.LITTLE technology \cite{armbiglittle}.

On the other hand, the Tegra 186 SoC relies on a heterogeneous architecture that combines two high-performance Armv8-compliant VLIW cores 
developed by NVIDIA (\emph{Denver 2}) with four out-of-order Arm Cortex A57 cores. Given that only the T124 SoC offered the possibility of 
running code at a true low-power CPU, we benchmarked the serial NMS code on the two CPU profiles available on that platform (i.e. 
high-performance CPU cores and the low-performance CPU core). Referring again to the GPU capabilities, even though that both T210 and T186 SoCs feature 256 CUDA cores, the latter one 
both runs at a higher clock rate and doubles the effective GPU memory bandwidth by increasing the memory bus from 64 bits to 128 bits. This fact 
will enable us later to study the impact of doubling the memory bandwidth when executing on the GPU the parallel \emph{map/reduce} NMS kernels.

 \begin{table}
  \renewcommand{\arraystretch}{1.3}
  \caption{Tested embedded boards, Linux distributions, and CUDA runtime versions.}
  \centering
  \footnotesize
  \begin{tabular}{| c | c | c | c | c | c |}
  \hline
  \textbf{Board} & \textbf{Tegra SoC} & \textbf{JetPack} & \textbf{L4T} & \textbf{CUDA} \\
  \hline
  Jetson TK1 & T124 & v1.2 & R21.5 & 6.5 \\
  Jetson TX1 & T210 & v3.2.1 & R28.2 & 9.0 \\
  Jetson TX2 & T186 & v3.2.1 & R28.2 & 9.0 \\
  \hline
  \end{tabular}
  \normalsize
  \label{tab:jetsondistros}
 \end{table}

 \begin{table}
  \renewcommand{\arraystretch}{1.3}
  \caption{\texttt{sysfs} path for setting up GPU clock rates.}
  \centering
  \footnotesize
  \begin{tabular}{| c | c |}
  \hline
  \textbf{Tegra SoC} & \textbf{Filesystem Path} \\
  \hline
  T124 & \texttt{/sys/kernel/debug/clock/gbus/} \\
  T210 & \texttt{/sys/devices/57000000.gpu/} \\
  T186 & \texttt{/sys/devices/17000000.gp10b/} \\
  \hline
  \end{tabular}
  \normalsize
  \label{tab:tegrafs}
 \end{table}

Finally, the most important part of the evaluation was to carefully select the input datasets to demonstrate how the proposed parallel NMS 
is able to cope with challenging real-world situations that feature lots of simultaneous objects in high-resolution images and video frames. 
As we are interested in human faces, we selected a 1080p input video from the 
\emph{SVT High Definition Multi Format Test Set} \footnote{\url{ftp://vqeg.its.bldrdoc.gov/HDTV/SVT_MultiFormat/}} (named 
\texttt{crowd\_run}), which was originally filmed by the Swedish Television at 50 FPS using 65 mm film professional equipment. This particular 
video features approximately on average 60 simultaneous faces, but it was especially chosen because it yields hundreds of simultaneous detections per frame after 
having executed a CNN-based face classifier.

\begin{figure}[t]
  \centering
  \includegraphics[scale=0.25]{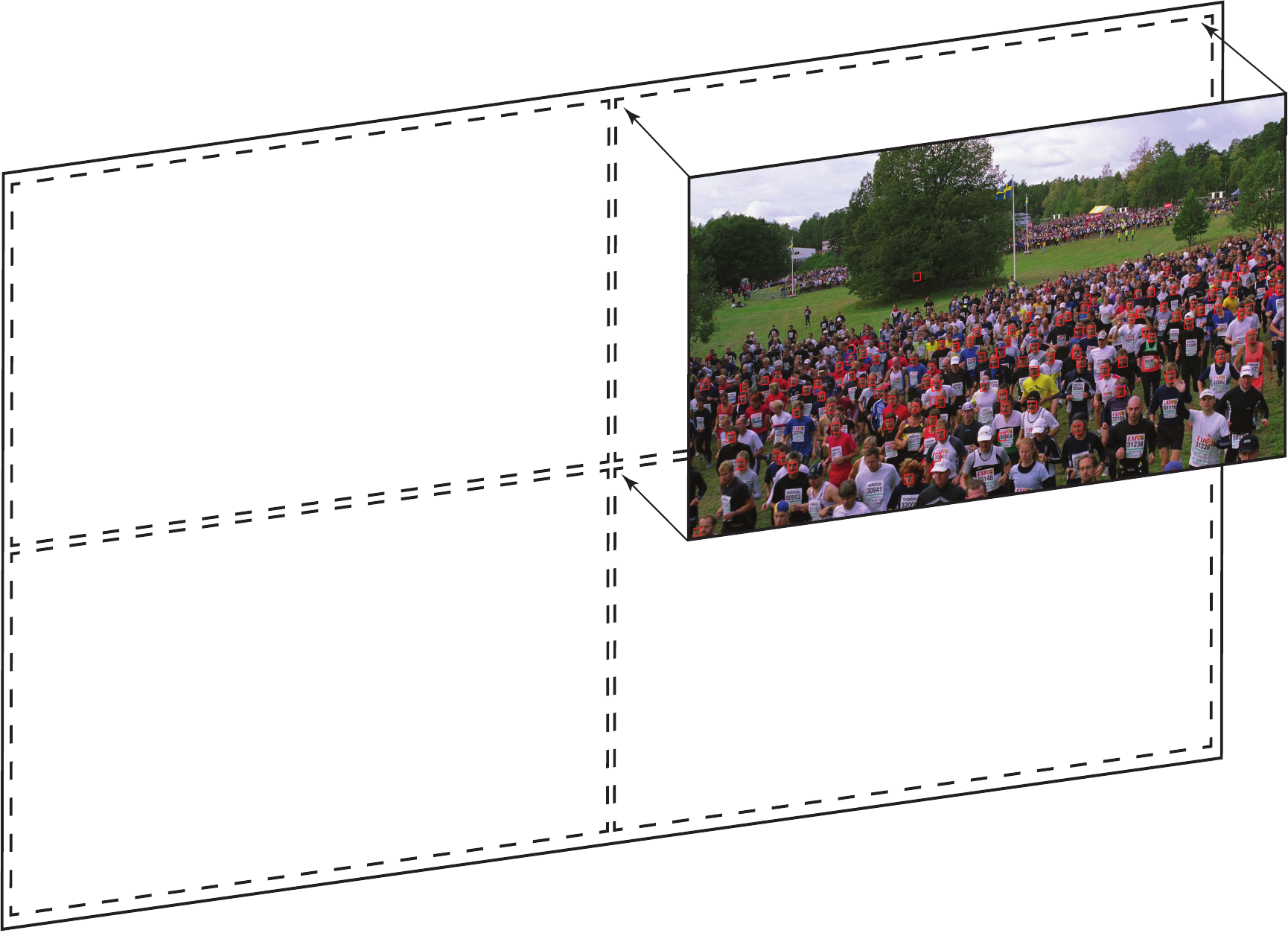}
  \caption{\label{fig:nmsmosaic} Synthetic 3840x2160 video frame from the \texttt{mosaic} dataset (post-NMS).}
\end{figure}

In order to further stress out the parallel NMS kernels, we post-processed the original 1080p \texttt{crowd\_run} video stream to generate a 3960x2160 synthetic 
\texttt{mosaic} video incorporating four streams on each frame (depicted in Figure \ref{fig:nmsmosaic}), thus effectively quadrupling the number of simultaneous 
detected faces. Similarly, we also decided to study in detail the scalability of parallel NMS as the number of simultaneous faces are linearly increased. For this 
particular experiment, we selected a picture of the \emph{83rd Academy Awards Nominees} \footnote{\url{http://tinyurl.com/o5n97ra/}} captured during 
the Oscars ceremony (named \texttt{oscars} in Table \ref{tab:nmsinputdatasets}), which shows 147 simultaneous 
faces. Once all faces were detected, we coded and executed a script that generated 147 synthetic images from the original picture by covering and uncovering faces 
using rectangles filled with black color (see Figure \ref{fig:nmsoscars}). Therefore, a given $p_i$ picture (where $1 \leq i \leq 147$) would just uncover an additional 
face when compared to the previous $p_{i-1}$ one. As an example, picture $p_{0}$ starts the process by having all localized faces (147) covered with black 
rectangles to enforce the face classifier to yield zero detections. Under this rationale, picture $p_{1}$ would cover 146 faces with black 
rectangles and uncover only a single face throughout the whole picture. As the script keeps going on generating synthetic images, picture $p_{147}$ 
would conclude the process by showing all faces completely uncovered.

\section{Obtained Results}

\begin{figure}[!t]
  \centering
  \includegraphics[scale=0.33]{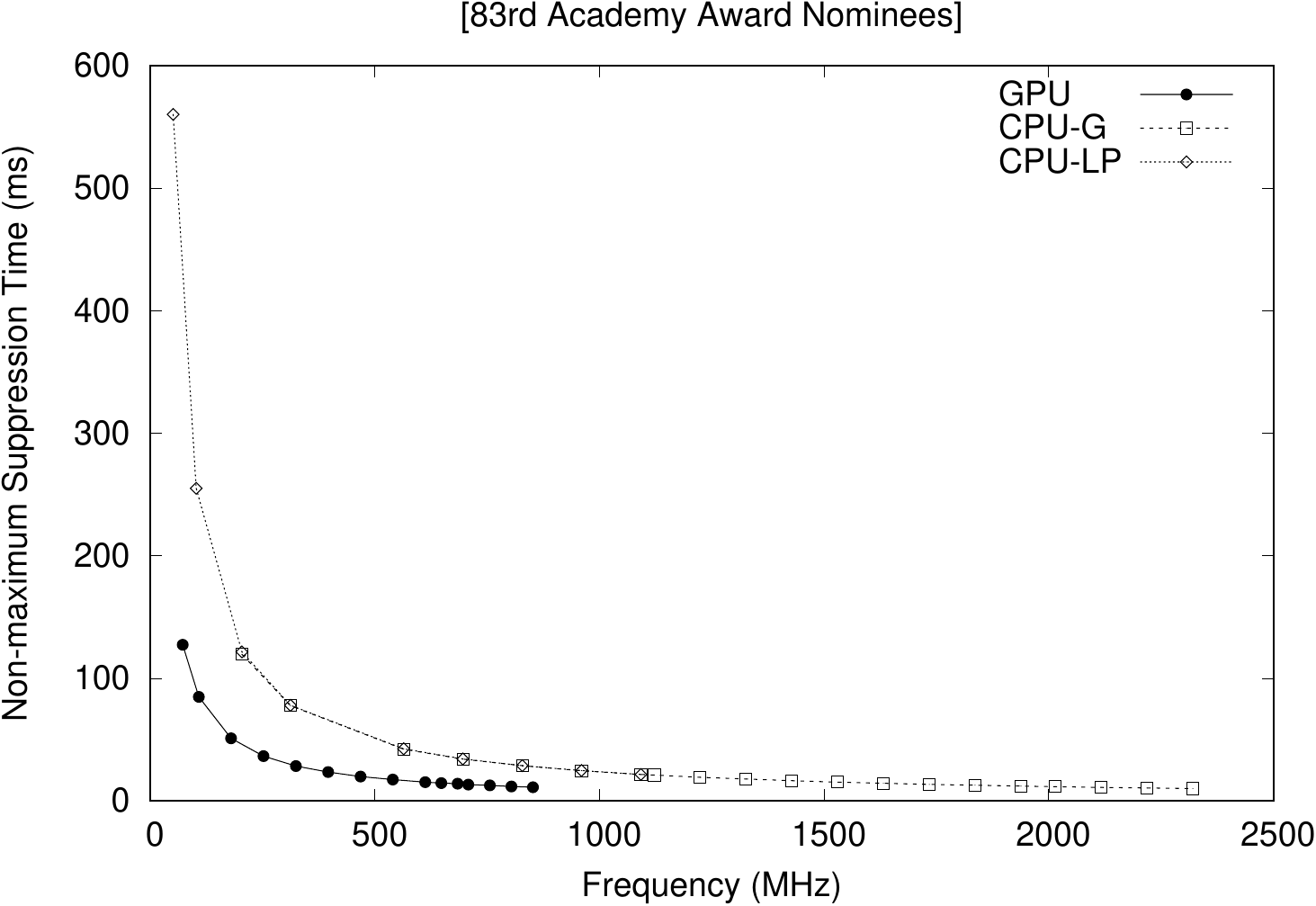}
  \caption{\label{fig:nmscpugpufreq} NMS latency on GPU, CPU-LP and CPU-G cores for the selected \texttt{oscars} dataset.}
\end{figure}

As it was previously discussed, first we started by determining how the latency of the GPU-based parallel NMS algorithm compares 
against the traditional serialized version of OpenCV's $\mathcal{O}(n^{2})$ greedy NMS, which is still used in many 
works \cite{huval2015empirical} for clustering the bounding boxes obtained after inferencing CNN classifiers. This comparison 
was conducted by executing both NMS algorithms on the Jetson TK1 board using as an input the \texttt{oscars} dataset picture. 
More precisely, both NMS algorithms were executed at multiple frequencies on the two different CPU clusters available in the T124 SoC 
(i.e. high-performance CPU-G cluster, and low-power CPU-LP cluster) by properly setting up the parameters in the \texttt{sysfs} Linux kernel interface. 
The CPU-LP core clock frequency ranges between 51 MHz and 1 GHz, while a given CPU-G core ranges between 204 MHz and 2.3 GHz. Similarly, 
the GPU clock ranges between 72 MHz and 850 MHz, as Table \ref{tab:jetsonplatforms} shows. The results of this experiment are shown in 
Figure \ref{fig:nmscpugpufreq}.

\begin{table}
  \renewcommand{\arraystretch}{1.3}
  \caption{Input datasets selected for benchmarking the parallel NMS kernels.}
  \centering
  \footnotesize
  \begin{tabular}{| c | c | c |}
  \hline
  \textbf{Input Dataset} & \textbf{Resolution} & \textbf{Type}  \\
  \hline
   \texttt{crowd\_run} & 1920x1080 & H.264 Video @ 50 FPS \\
   \texttt{mosaic} & 3840x2160 & H.264 Video @ 50 FPS \\
   \texttt{oscars} & 4646x1800 & JPEG Picture \\
  \hline
  \end{tabular}
  \normalsize
  \label{tab:nmsinputdatasets}
 \end{table}

The graph clearly shows that a GPU clocked at 50\% of its maximum frequency outperforms both CPU types 
also when they are operating at 50\% of its peak operating frequency. For this experiment, the overhead of memory allocations 
and initializations was not taken into account, and only the pure kernel execution time was considered. It should be noted 
that for battery-powered fanless solutions the chip must be underclocked or automatically managed by the underlying DVFS 
subsystem to avoid excessive overheating. 

\begin{figure}[h]
  \centering
  \setlength{\fboxsep}{0pt}\fbox{\includegraphics[scale=0.12]{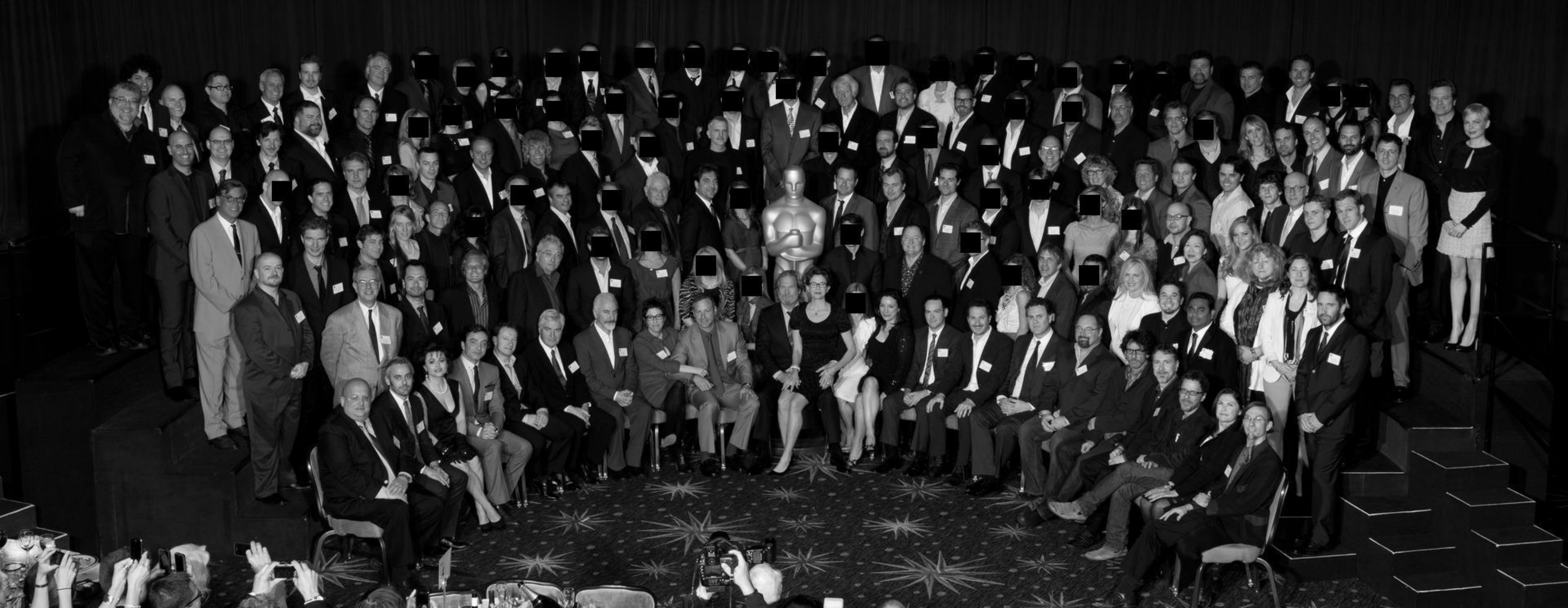}}
  \caption{\label{fig:nmsoscars} Synthetic \emph{83rd Academy Awards Nominees} picture showing part of the faces covered.}
\end{figure} 

Unsurprisingly, the GPU shows an advantage, especially for GPU-based object detection and 
recognition pipelines, in which unwanted CPU/GPU memory transfers slowdown the throughput of real-time applications, as it 
would undesirably happen when relying on the serialized NMS targeting CPUs.

\begin{figure}
  \centering
  \includegraphics[scale=0.33]{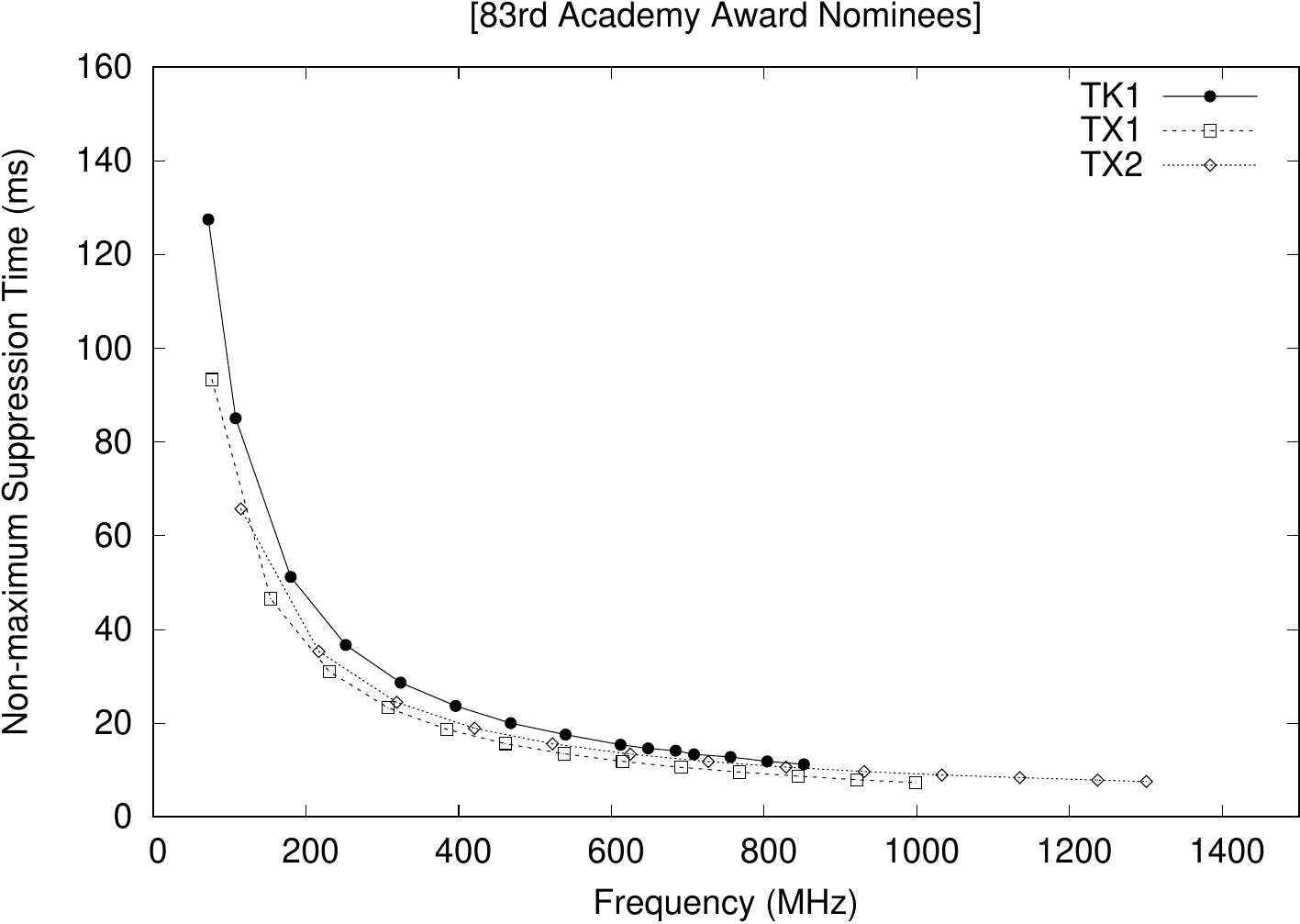}
  \caption{\label{fig:nmsgpufreq} NMS latency when processing the \texttt{oscars} dataset on the GPUs included in the selected Tegra SoCs.}
\end{figure}

On the other hand, the proposed parallel NMS was evaluated on all GPUs of the selected Tegra platforms, also by varying the clock rate of the GPU before executing  
the kernels during the experiments, and using the same input dataset (\texttt{oscars}). The results of these experiments are 
summarized in Figure \ref{fig:nmsgpufreq}. The reported latencies, aggregate the combined execution time of the \emph{map/reduce} 
kernels within the parallel NMS algorithm, and shows a latency reduction improvement when scaling from 192 CUDA cores (Tegra T124) 
to 256 (Tegra T186 and T210). More precisely, the latencies obtained when running the kernels at the maximum GPU clock 
frequency were 11.24 ms (T124), 7.36 ms (T210), and 7.6 ms (T186), respectively, when clustering roughly 3000 simultaneous 
detections. Unexpectedly, although each experiment was executed three times to avoid the bias introduced by the CUDA runtime 
and scheduler, the obtained NMS algorithm performance was slightly faster on the Jetson TX1 board (T210) when compared to 
the results gathered on the Jetson TX2 (T186), which features a wider memory bus.

The obtained latency figures did not prove the general assumption that executing kernels on GPUs with equal core counts, but 
which differ in memory bandwidth and clock frequencies, would yield better performance results when physical specifications 
are improved across chip generations (see Table \ref{tab:jetsonplatforms}). Quite the contrary, it showed that for our   
proposed parallel NMS algorithm, it seems far more important to increase by a 33\% the amount of GPU cores rather than 
simply increase the clock frequency and memory bandwidth, given a constant number of GPU cores. As an example, the T186 GPU 
must be clocked at 1.3 GHz to achieve the performance figures yielded by the T210 chip when clocked at 998.4 MHz. Therefore, 
the GPU scalability of parallel NMS is more sensitive to the number of cores (a 33\% increase in GPU cores 
yields a 55\% latency reduction), rather than to the improvements of the memory subsystem, thus highlighting that kernel code 
bottlenecks are skewed towards the availability and quantity of ALUs included in the GPUs (i.e. compute-bound kernel).

\begin{table*}
\renewcommand{\arraystretch}{1.3}
\scriptsize
\centering
\begin{tabular}{|c|c|c|c|c|c|c|}
\hline
\textbf{Discrete GPU} & \textbf{CUDA Cores} & \textbf{Memory Type} & \textbf{Clock Rate} & \multicolumn{3}{c|}{\textbf{NMS Latency (ms)}} \\
\hline
 & & & & $n = 200$ & $n = 1027$ & $n = 2895$ \\ \cline{5-7}
GeForce GTX 1060 & 1280 & GDDR5 & 1.70 GHz & 0.107 & 0.324 & 1.313 \\
Tesla T4 & 2560 & GDDR6 & 1.59 GHz & 0.024 & 0.088 & 0.430 \\
\hline
\end{tabular}
\caption{NMS latency on discrete GPUs when clustering \emph{n} detections (\texttt{oscars} dataset).}
\label{tab:nmsdiscretegpu}
\end{table*}

\begin{figure}[t]
  \centering
  \includegraphics[scale=0.35]{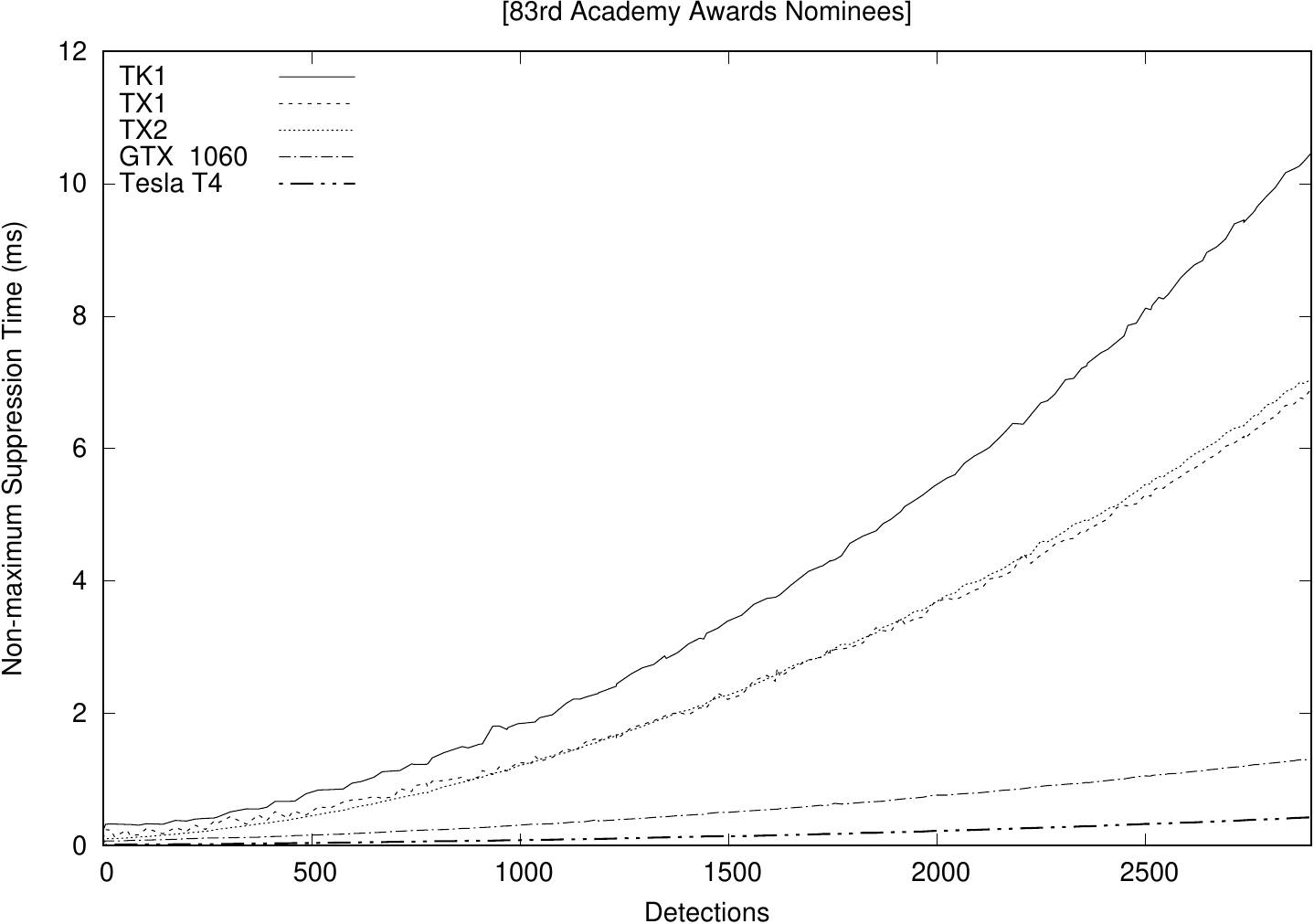}
  \caption{\label{fig:nmsdetectionsscaling} NMS latency on growing number of detections (\texttt{oscars} dataset).}
\end{figure}

\begin{figure*}[!t]
  \centering
  \includegraphics[scale=0.7]{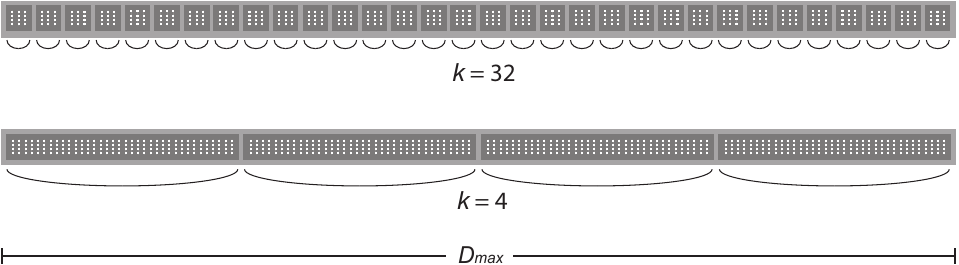}
  \caption{\label{fig:kparametercomp} Thread synchronizations computed on a given B matrix row (\emph{k=32} and \emph{k=4}).}
\end{figure*}

However, since the main benefit of the parallel NMS is the potential and flexibility offered for handling the 
clustering of detections in workloads featuring huge amounts of simultaneous objects, we decided to 
determine the scalability of the \emph{map/reduce} kernels as the quantity of simultaneous faces are increased. These 
experiments involved the execution of the parallel NMS kernels on the selected Tegra platforms using as an input the synthetic 
set of pictures featuring covered faces (as it is shown in Figure \ref{fig:nmsoscars}). The results of such experiments 
are summarized in the graph depicted in Figure \ref{fig:nmsdetectionsscaling}, and are consistent with the previous observation 
that emphasizes the importance of the GPU CUDA core count (both TX1 and TX2 outperform the older TK1). Hence, the slopes 
of both TX1 and TX2 graphs are greatly reduced when compared to the baseline TK1 graph, thus proving that our proposed parallel 
NMS algorithm properly scales as the CUDA core count is increased. In order to further illustrate this fact, Figure 
\ref{fig:nmsdetectionsscaling} also shows an extra reduction of the graph slope when the parallel NMS algorithm is executed on 
a discrete NVIDIA GeForce GTX 1060 GPU featuring 1280 CUDA cores over a growing number of detections, and a further reduction 
when executed on a Tesla T4 GPU with 2560 CUDA cores. More particularly, the GTX 1060 yielded an 8.11X speed up over the TK1, 
5.38X (TX2) and 5.23X (TX1), respectively. And the Tesla T4 yielded a 24.59X speed up over the TK1, 15.91X over TX2, and 
16.44X over TX1.

As Table \ref{tab:nmsdiscretegpu} shows, discrete GPUs clocked at roughly the same frequencies, and relying on comparable memory
subsystem technologies (i.e. GDDR5 -GTX 1060- vs. GDDR6 -Tesla T4-), but with increased number of CUDA cores trigger substantial
reductions in the NMS latency. As a result, increasing by a factor of 2 the amount of cores in the underlying hardware resources
reduces the NMS execution latency by a factor of 3 for 2895 detections, effectively proving the scalability of our parallel NMS
method as the amount of cores keeps increasing.

These results are in line with our initial expectations about the time complexity of the parallel NMS. On these latter experiments, the GPU kernels 
were also compiled and benchmarked by setting $k=32$ and $D_{max}=4096$ (upper limit used for the GPU memory allocations required 
by matrix $B$ and vector $V$). 

\begin{figure}[t]
  \centering
  \includegraphics[scale=0.33]{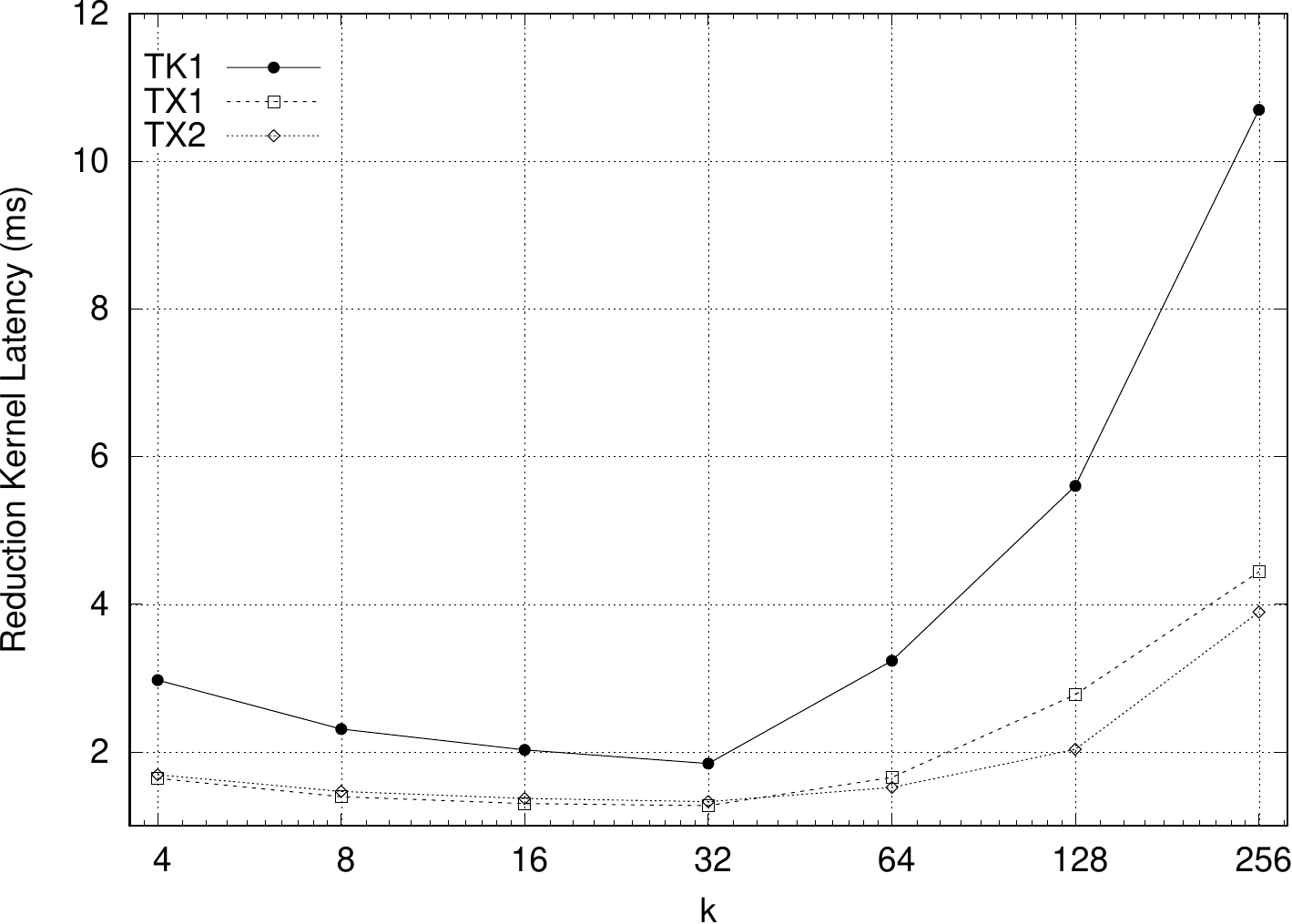}
  \caption{\label{fig:nmskparametergraph} NMS reduction kernel latency according to parameter \emph{k}.}
\end{figure}

In order to further study and optimize the NMS reduction kernel (Algorithm \ref{alg:reduce}), several experiments were carried 
out to study the impact of parameter $k$ with the aim of finding which partition size minimizes the latency in the NMS 
reduction kernel. Figure \ref{fig:nmskparametergraph} summarizes the obtained results on each SoC of the selected 
Tegra platforms on the \texttt{oscars} dataset.

According to the experiments, the optimal partition size yielding minimal latencies would be $k = 32$, which was precisely the value used for 
obtaining the graph shown in Figure \ref{fig:nmsdetectionsscaling}. Interestingly, these results challenge 
conventional wisdom, as it is increasingly less costly to call $k$ times \small\texttt{\_\_syncthreads\_and() }\normalsize 
 in the \textbf{for} loop when $4 \leq k \leq 32$. As thread synchronization is typically slow (i.e. theoretically, a barrier should be avoided 
if possible), it is considered counter-intuitive that an increasing number of thread synchronization operations could contribute to reduce the 
execution time. Taking into account that NVIDIA GPUs offer architectural support at the ISA level for implementing these primitives, and GPU hardware thread 
synchronization primitives work at the CUDA block level, the amount of $D_{max} / k$ partitions created per matrix row effectively serves as a 
mechanism for fine-tuning the quantity of inter-block \small\texttt{\_\_syncthreads\_and() }\normalsize calls executed in parallel (represented 
as arc symbols in Figure \ref{fig:kparametercomp}). A comparison of the granularity of such thread synchronizations is highlighted in the  
aforementioned figure, which illustrates how the \emph{k} parameter greatly affects the number of \small\texttt{\_\_syncthreads\_and() }\normalsize 
calls executed in parallel, as the amount of CUDA blocks created varies. Nevertheless, when $k > 32$, as Figure \ref{fig:nmskparametergraph} depicts, 
the NMS reduction kernel latency keeps growing due to unoptimal CUDA block partitions when processing the input matrix $B$.

\begin{figure}[t]
  \centering
  \includegraphics[scale=0.33]{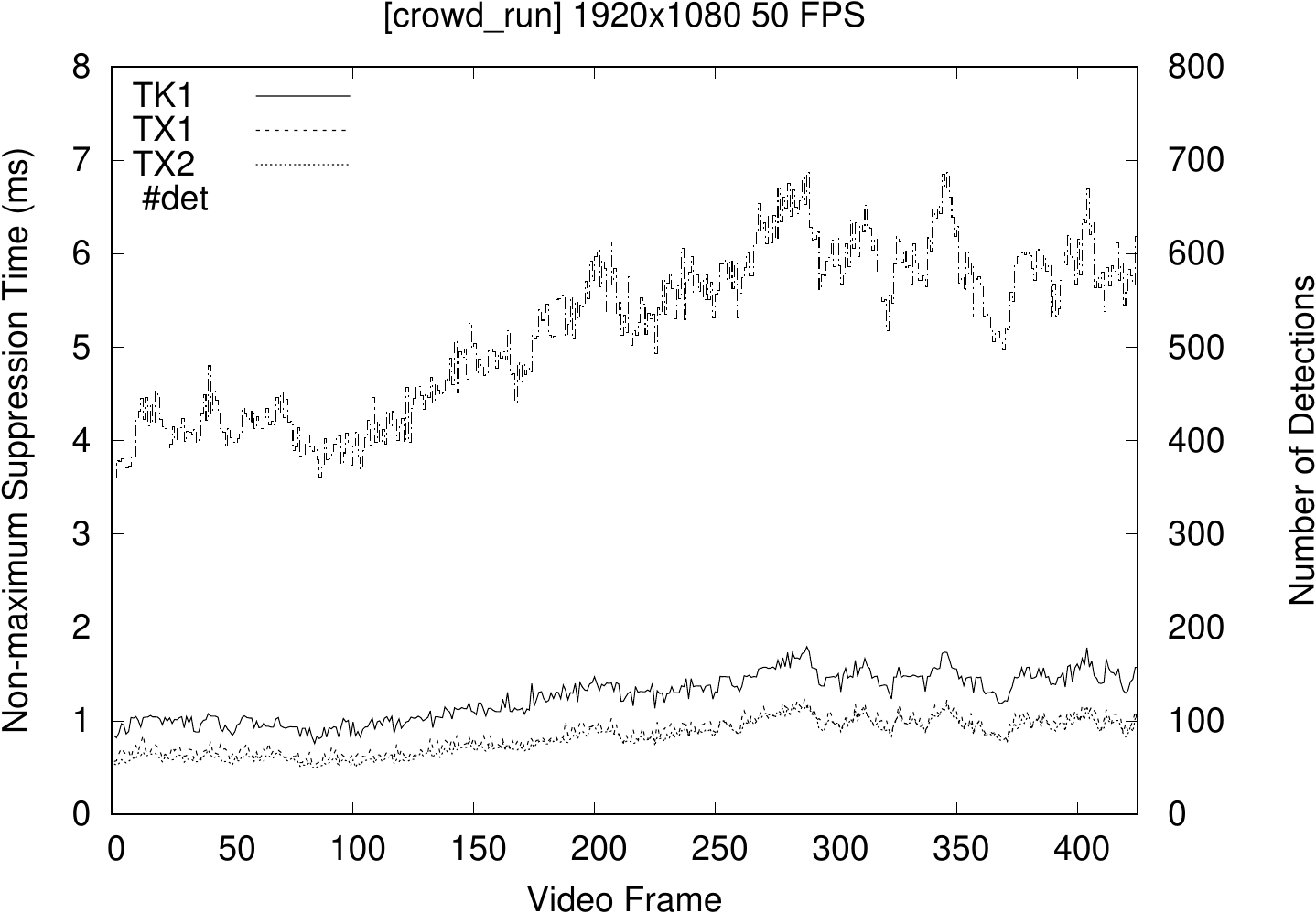}
  \caption{\label{fig:crowdrungraph} NMS latency per frame vs. \texttt{\#det} number of detections (\texttt{crowd\_run}).}
\end{figure} 

Finally, the \texttt{crowd\_run} and \texttt{mosaic} datasets described in Table \ref{tab:nmsinputdatasets} were also used as an input of the 
\emph{map/reduce} NMS kernels. The main objective of experimenting with those datasets was to characterize the underlying performance scalability 
of both kernels when localizing faces in videos featuring challenging real-world scenarios. Additionally, further experimentation was carried out 
to determine the execution time distribution of both Algorithm \ref{alg:map} (\emph{map} kernel) and Algorithm \ref{alg:reduce} (\emph{reduce} kernel).

\begin{figure}[t]
  \centering
  \includegraphics[scale=0.33]{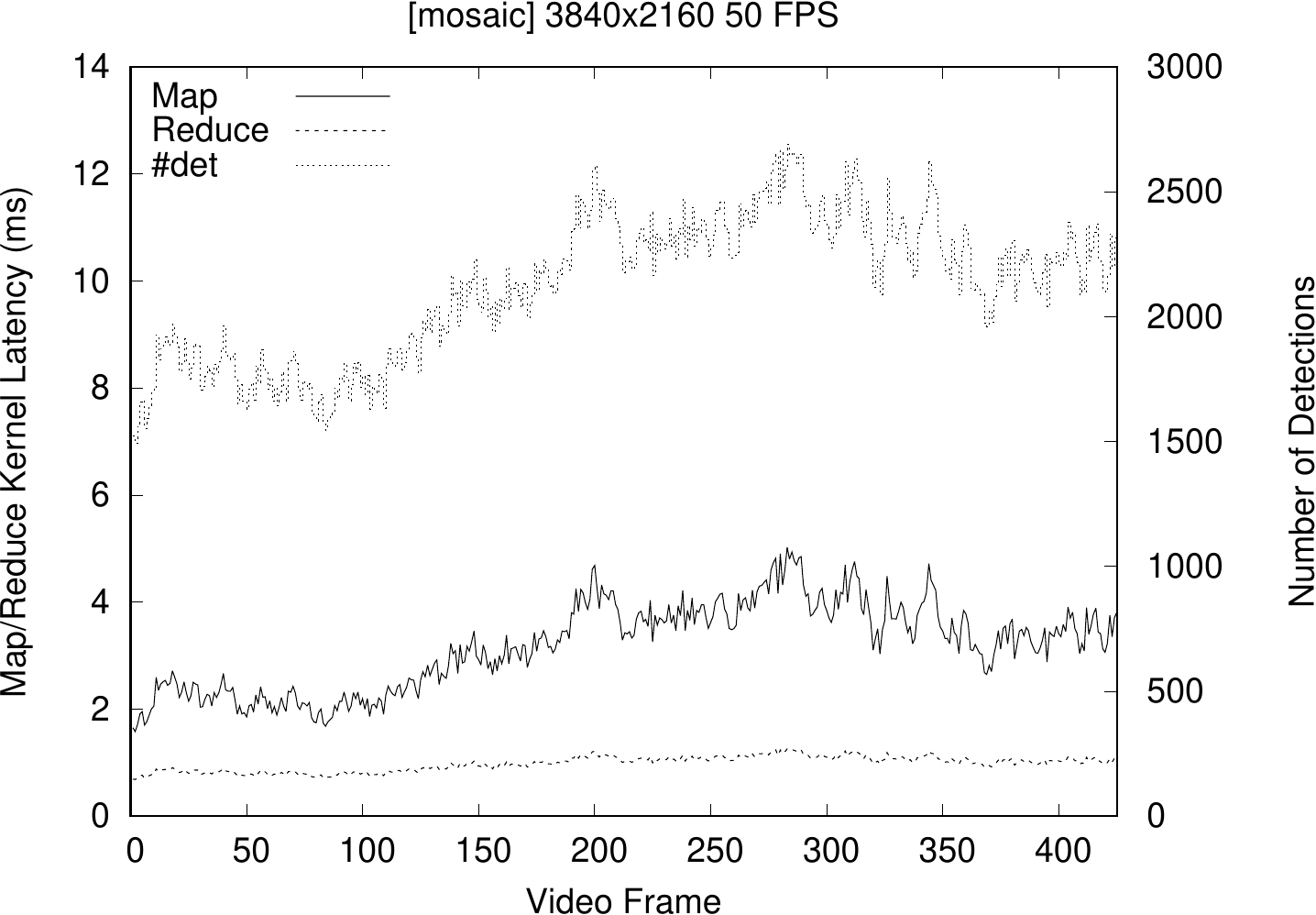}
  \caption{\label{fig:mosaicmapreduce} Map/Reduce kernel latency per frame vs. \texttt{\#det} number of detections (\texttt{mosaic}).}
\end{figure} 

 \begin{table}
  \renewcommand{\arraystretch}{1.3}
  \caption{Average NMS latencies on TK1, TX1 and TX2 SoCs (\texttt{mosaic} dataset).}
  \centering
  \footnotesize
  \begin{tabular}{| c | c | c | c | c |}
  \hline
  \multicolumn{5}{|c|}{\textbf{Average NMS Latency (ms)}} \\
  \hline
    Tegra SoC & \emph{k} & Map & Reduce & Total \\
  \hline
    \multirow{4}{*}{TK1} & 4  & 5.17 & 2.41 & 7.78 \\
                         & 8  & 5.16 & 1.87 & 7.22 \\
                         & 16 & 5.17 & 1.64 & 6.97 \\
                         & 32 & 5.17 & 1.54 & 6.88 \\
  \hline
    \multirow{4}{*}{TX1} & 4  & 3.74 & 1.34 & 4.84 \\
                         & 8  & 3.37 & 1.11 & 4.57 \\
                         & 16 & 3.38 & 1.03 & 4.53 \\
                         & 32 & 3.38 & 1.00 & 4.48 \\
  \hline
    \multirow{4}{*}{TX2} & 4  & 3.52 & 1.37 & 4.93 \\
                         & 8  & 3.50 & 1.16 & 4.71 \\
                         & 16 & 3.50 & 1.07 & 4.61 \\
                         & 32 & 3.49 & 1.03 & 4.56 \\
  \hline
  \end{tabular}
  \normalsize
  \label{tab:nmsmosaicmean}
 \end{table}

As it was done with the \texttt{oscars} dataset picture, the NMS kernels were benchmarked over time on the selected Tegra platforms 
after having firstly decoded H.264 video frames, and later performed face detection. The obtained results are summarized in Figure \ref{fig:crowdrungraph}, 
which shows both the number of faces detected per frame (top dashed line), and the aggregated latency of the parallel NMS kernels when 
executed on the selected SoCs (three lines shown at the bottom). Therefore, the H.264 decoding latency was not taken into account thus 
profiling only the \emph{map/reduce} kernels at the low level.

On this particular dataset (\texttt{crowd\_run}), the amount of detected faces per frame ranged between 360 and 698 detections. Again, these experiments 
were carried out by setting $k=32$ and $D_{max}=4096$ when clustering detected faces. The aggregated latencies yielded by the parallel NMS were 
less than 2 ms. More particularly, it ranged between 0.5 ms and 1.8 ms when executed on the Jetson TK1 (T124). When switching to Jetson TX1 (T210) and 
Jetson TX2 (T186) platforms, latencies were reduced a further 50\% due to the increased GPU core counts. As it also happened with previous experiments, there 
were no major performance improvements between TX1 and TX2 platforms when executing the parallel NMS algorithm. The latency reduction on TX2 was on 
average 6.25\% lower than on TX1 and very close to 0.5 ms. 

Therefore, in order to further spot differences and analyze the potential benefits of the improved T186 SoC, it might be necessary to 
saturate the GPU resources. Moreover, the \texttt{mosaic} video dataset was used as an input to quadruple the number of simultaneous detections 
per frame, so that the amount of thread reductions per row triggered by the NMS reduction kernel is greater than one. The results of 
these experiments are summarized in Table \ref{tab:nmsmosaicmean}, which details the average latencies for both \emph{map} and \emph{reduce} kernels 
after having executed them on a frame per frame basis. Additionally, Table \ref{tab:nmsmosaicmean} reports how the execution time of the parallel NMS kernels 
was affected by parameter $k$ on the selected Tegra platforms. 

According to the obtained measurements, after quadrupling the amount of simultaneous faces, the main bottleneck of the parallel 
NMS algorithm lies on the construction of matrix $B$, which is performed in Algorithm \ref{alg:map} (\emph{map} kernel). This 
performance penalty is noticeable in Figure \ref{fig:mosaicmapreduce}, as the depicted \emph{map} kernel latency per frame seems  
proportional to the number of detected faces. These results are in line with the inner workings of Algorithm \ref{alg:map} since 
this kernel must populate $n^2$ elements to cluster $n$ detections. Consequently, it is the \emph{map} kernel the one saturating 
the underlying hardware resources. On the other hand, the \emph{reduce} kernel latency per frame remains roughly constant throughout 
the \texttt{mosaic} video frames, even when the number of simultaneous faces detected per frame is dramatically increased.

 \begin{table}
 \caption{Comparison between our parallel greedy NMS method and other state-of-the-art NMS methods on the PASCAL VOC2007 and CrowdHuman datasets.}
 \centering
 \footnotesize
 \begin{tabular}{cc}
  \begin{tabular}[t]{| c | c |}
   \hline 
   \multicolumn{2}{|c|}{\textbf{VOC2007 (mAP)}} \\
   \hline
   \emph{This work} & 0.7516 \\
   NMSNet~\cite{qiu2019graph} & 0.7659 \\
   HS-NMS~\cite{song2019improved} & 0.7556 \\
   Soft-NMS~\cite{bodla2017soft} & 0.7623 \\
   \hline   
  \end{tabular}
  \hfill
  \begin{tabular}[t]{| c | c |}
   \hline 
   \multicolumn{2}{|c|}{\textbf{CrowdHuman (AP)}} \\
   \hline
   \emph{This work} & 0.8307 \\
   AdaptiveNMS~\cite{liu2019adaptive} & 0.8471 \\
   FeatureNMS~\cite{salscheider2020featurenms} & 0.8538 \\
   Soft-NMS~\cite{bodla2017soft} & 0.8392 \\
   \hline   
  \end{tabular}
 \end{tabular}
 \normalsize
 \label{tab:accuracycomp}
 \end{table}  

To conclude, we compared the accuracy of our parallel NMS method relying on the classic greedy approach to the recent NMS methods mentioned in Section \ref{sec:rwork}. 
This comparison was conducted over the PASCAL VOC2007 and CrowdHuman datasets. For the algorithms lacking publicly available source code, we used the mAP and AP figures 
as reported by their respective authors when evaluating both datasets. The obtained results are summarized in Table \ref{tab:accuracycomp}, which clearly shows that 
alternative state-of-the-art NMS methods only marginal improve (between 1\% and 2\% in terms of mAP and AP depending on the dataset) the accuracy of our parallel NMS method. 
Moreover, alternative sequential methods applying minor variations to the classic greedy approch (i.e. AdaptiveNMS, Soft-NMS, and FeatureNMS) could also be potentially 
adapted to GPU architectures using the parallelization techniques presented in Section \ref{sec:impl}.

\section{Algorithm Correctness}

\textsc{MapKernel($B$, $D$)} specification:\\

\noindent \textbf{Precondition:} $D$ is a vector of unsorted detections of size $D_{max}$, and $B[0...n-1][0...n-1]$ is a $n \times n$ boolean 
matrix in which all elements are set to \emph{true} (represented with value 1). The dimensions of matrix $B$ correspond to the 
size of vector $D$ ($n = D_{max}$). \\

\noindent \textbf{Postcondition:} $\forall i,j: 0 \leq i \leq n-1, 0 \leq j \leq n-1:$ A given $B_{ij}$ element in boolean matrix $B$ encode if 
$A(d_{i} \cap d_{j}) / A(d_{j}) < \theta$ and \emph{score($d_{i}$)} $<$ \emph{score($d_{j}$)}, in which detections $d_{i},d_{j} \in D$, and 
$n = D_{max}$. \\

\begin{theorem}
The CUDA kernel pseudocode of Algorithm \ref{alg:map} meets the \textsc{MapKernel($B$, $D$)} specification shown above, in which 
all threads created independently generate in parallel a single element of boolean matrix $B$.
\end{theorem}

\begin{proof}
The input matrix $B$ values are generated in parallel by creating $D_{max} \times D_{max}$ threads using CUDA 2D block partitions of size 
\footnotesize\texttt{blockDim * blockDim}\normalsize. This two-dimensional parallelization pattern ensures that a given $(i,j)$ thread 
generates only a single $B_{ij}$ boolean matrix element in the intervals $0 \leq i \leq n-1$ and $0 \leq j \leq n-1$. As $i,j$ indexes 
detections $d_{i}$ and $d_{j}$, respectively; and the $D$ vector is only accessed by threads within the kernel for read operations (i.e. never written), 
while $B$ matrix values are never read by a thread, there is no chance for experiencing race conditions. Therefore, it is guaranteed that a 
given $(i,j)$ thread will independently produce a single $B_{ij}$ element as all \emph{D} data dependencies are read-only and remain unmodified 
throughout the whole kernel execution. The $i$ variable used in the kernel indexes $B$ rows, whereas $j$ is used to index the matrix columns. 
Hence, the kernel overwrites a given $B_{ij}$ element only when the score of detection $d_j$ exceeds the score of detection $d_i$. Due to the 
abovementioned 2D parallelization pattern, when the kernel execution concludes, all possible detection pairs ($d_i,d_j$) derived from vector $D$ will be 
compared against each other. On the other hand, $w$ and $h$ variables are set to the maximum width and height window dimensions of the 
considered ($d_i,d_j$) detection pair for computing the thresholded clipping process with $\theta$ constant. As a result of this, the value 
stored in $B_{ij}$ will be set with the boolean obtained by evaluating the thresholded area intersection $A(d_{i} \cap d_{j}) / A(d_{j}) < \theta$. 
Therefore, after concluding the execution of the parallel kernel it is guaranteed that the postcondition is correctly met.
\end{proof}

\bigbreak

\noindent \textsc{ReduceKernel($B$, $k$, $V$)} specification:\\

\noindent \textbf{Precondition:} $B$ is $n \times n$ boolean matrix obtained after the execution of \textsc{MapKernel}, and $V$ a 
boolean vector o size $n$ ($n = D_{max}$) in which all elements have been set to \emph{true}. The $k$ value is a parameter used 
to fine-tune the size of row partitions of matrix $B$ when performing computations during kernel execution. \\

\noindent \textbf{Postcondition:} $\forall i: 0 \leq i \leq n-1: V_i = \bigwedge_{j=0}^{n-1} (B_{ij})$ where $n = D_{max}$. \\

\begin{theorem}
The kernel pseudocode of Algorithm \ref{alg:reduce} populates vector $V$ by meeting the postcondition of 
\textsc{ReduceKernel($B$, $k$, $V$)} described in the specification. The precondition must be satisfied by 
executing such kernel immediately after \textsc{MapKernel($B$, $D$)}.
\end{theorem}

\begin{proof}
The unidimensional parallelization pattern computes a given $V_i$ element of vector $V$ by creating $k$ subsets of $D_{max} / k$ 
threads (i.e. CUDA block size of dimensions $(D_{max} / k) \times 1$), where $0 \leq i \leq D_{max}$. Accordingly, the 
statement $i \leftarrow $ \footnotesize\texttt{blockIdx.x} \normalsize ensures that a given CUDA thread block is mapped 
to the $i$-th row of a $B^{k}$ submatrix of $B$ containing $D_{max} \times (D_{max} / k)$ boolean elements. By following this scheme, 
$k$ submatrices of input $B$ are considered, where $B^{k} \subseteq B$:  

\[
B^{k} =
  \begin{bmatrix}
    b_{00} & b_{01} & \cdots & b_{0\left\lfloor\frac{D_{max}}{k}\right\rfloor-1} \\
    b_{10} & b_{11} & \cdots & b_{1\left\lfloor\frac{D_{max}}{k}\right\rfloor-1} \\
    \vdots & \vdots & \ddots & \vdots \\
    b_{D_{max-1}0} & b_{D_{max-1}1} & \cdots & b_{D_{max-1}\left\lfloor\frac{D_{max}}{k}\right\rfloor-1}
  \end{bmatrix} 
\] \\

\noindent Therefore, in total, $D_{max}$ CUDA blocks of dimensions $(D_{max} / k) \times 1$ are created, in which 
the $j$-th thread in the block is mapped to access in parallel a different row element of the submatrix, 
where $0 \leq j \leq D_{max} / k$. As a result of this, the $V[i] \leftarrow $ \footnotesize\texttt{\_\_syncthreads\_and(B[j]) }\normalsize 
statement performs the aggregated AND operation of all elements on a given $i$-th row of $B^k$ and stores results 
in each $V_i \in V$. At this point, vector $V$ contains only the aggregated AND of submatrix $B^0$, and thus still lacks to 
take into account the remaining $k-1$ submatrices of $B$ (i.e. $B^1,...,B^{k-1}$). These remaining AND operations are 
performed in the \textbf{for} 1 \textbf{to} $k-1$ \textbf{do} loop. Since the binary AND operation satisfies both the associative 
and commutative properties, the order in which AND operations are conducted between the different submatrices is completely 
irrelevant. Therefore, the postcondition is met simply by computing the aggregated AND operation of each $B^1, ..., B^{k-1}$ submatrix, 
with the partial AND results stored in $V_{i}$. These operations are performed in the kernel by executing the statement  
$V[i] \leftarrow $ \footnotesize\texttt{\_\_syncthreads\_and}\normalsize$(V[i]$ \&\& $B[j])$, where the inter-AND operations between 
the $B^0, ..., B^{k-1}$ submatrices and the partial AND results stored in $V_i$ are denoted by the $\&\&$ operand. Finally, 
when the execution of the kernel concludes, the boolean values stored in vector $V$ meets the expression shown in the postcondition, thus 
guaranteeing that each $V_i \in V$, contains a boolean value representing the aggregated AND of each row of the input $B$ matrix.
\end{proof}

\section{Conclusions}

In this paper, we have presented a novel highly scalable parallel NMS algorithm that is designed from scratch to handle 
workloads featuring thousands of simultaneous detections per frame. The proposed work-efficient NMS algorithm relies on a boolean 
matrix, which is constructed element wise in parallel, and completes the cluster of detections by means of parallel reductions. 
Additionally, the input set of candidate windows does not require to be pre-sorted before running our method, as the proposed 
map kernel always selects the representative with the highest score among all detections within the cluster while 
building the boolean matrix.

The obtained performance results show that the proposed \emph{map/reduce} kernels are compute bound, and properly scale on embedded 
GPUs as the amount of CUDA cores is increased. As a result of this, the parallel NMS algorithm is capable of clustering 1024 simultaneous 
detected faces per frame in roughly 1 ms on both Tegra X1 (T210) and Tegra X2 (T186) on-die GPUs, while taking 2 ms on Tegra K1 (T124). 
Therefore, the parallel NMS execution time is effectively reduced 53\% when the GPU computing resources are increased by a 33\% from 
192 to 256 CUDA cores. Thanks to additional experimentation, we also proved that this ratio is further improved as the amount 
simultaneous detections per frame keeps growing (e.g. 2048 detections and beyond).

Moreover, when the proposed parallel NMS method is executed on powerful discrete GPUs with high core counts and complex memory hierarchies, the 
execution time is even more drastically reduced. Interestingly, our obtained results show that doubling the amount of cores from 1280 to 2560 reduces the NMS 
execution time by at least a factor of three, which is even more pronounced than in the embedded GPU platforms. In this latter scenario, 
the parallel NMS is capable of clustering roughly 3000 candidate windows in less than 0.5 ms. These results show that our NMS method yields a minimal 
footprint in a GPU-only object recognition pipeline, as its execution consumes only 1\% of the hard deadline of the 40 ms required for the 
real-time processing of video feeds at 25 FPS.

On the other hand, the proposed kernels do not require to perform any GPU-to-CPU and CPU-to-GPU memory transfers, as the input of the \emph{map} kernel is directly 
populated with the output of a GPU-based object detection framework, thus involving only GPU-to-GPU memory transfers. Similarly, output data obtained after the execution 
of the \emph{map} kernel is directly fed to the input of the \emph{reduce} kernel within the GPU memory address space. This fact means that our parallel NMS method 
is the perfect solution for implementing an object recognition pipeline completely offloaded to GPU architectures.  

Regarding accuracy, as we do not modify the inner workings of the conventional sequential greedy NMS method, our parallel algorithm 
remains highly competitive \cite{bodla2017soft}. Furthermore, recent variations of the serial-based greedy NMS 
algorithm \cite{liu2019adaptive,salscheider2020featurenms} could also benefit from the scalability of the parallelization pattern and 
implementation studied in this paper.

Other alternatives, such as state-of-the-art learning-based methods, which require inferencing an additional CNN for solving the NMS problem, 
improve at most a 2\% \cite{hosang2017learning,qiu2019graph} the AP when compared to the classic NMS method, but 
dramatically increase both the computational burden and latency ($\sim$14x in \cite{hosang2017learning} and $\sim$40x in \cite{qiu2019graph}), which 
is key for real-time applications.

As future work, we plan to study the performance impact of implementing thread group synchronization in the reduction 
kernel grid. However, as this feature is only available on \emph{Pascal}, \emph{Volta}, and \emph{Ampere} GPU architectures, it would 
sacrifice portability of the parallel NMS kernel across legacy GPU computing platforms.

\section*{Acknowledgments}

This work has been partially supported by the Ministerio de Econom\'ia y Competitividad under contracts (TIN2015-65316-P, TEC2012-38939-C03-02), the 
Departament d'Innovaci\'o, Universitats i Empresa de la Generalitat de Catalunya under project MPEXPAR: Models de Programaci\'o i Entorns d'Execuci\'o 
Paral$\cdot$lels (2014-SGR-1051), and the European Commission under the Horizon 2020 program (H2020-ICT-644312).


\nocite{*}

\bibliographystyle{compj}
\bibliography{references}

\end{document}